\documentclass[sigconf]{acmart}

\setcopyright{none}
\copyrightyear{2018}
\acmYear{2018}
\acmDOI{XXXXXXX.XXXXXXX}
\acmConference[KDD '26]{the 32th ACM SIGKDD Conference on Knowledge Discovery and Data Mining}{August 9 to 13, 2026}{Jeju, Korea}
\acmISBN{978-1-4503-XXXX-X/2018/06}

\usepackage[utf8]{inputenc} 
\usepackage[T1]{fontenc}    
\usepackage{hyperref}       
\usepackage{url}            
\usepackage{booktabs}       
\usepackage{amsfonts}       
\usepackage{nicefrac}       
\usepackage{microtype}      
\usepackage{xcolor}         
\usepackage{adjustbox}
\usepackage{multirow}
\usepackage{boldline}
\usepackage{wrapfig}
\usepackage{amsmath}
\usepackage{graphicx}
\usepackage{bm}
\usepackage{caption}
\usepackage{subcaption}
\usepackage{marvosym}
\usepackage{colortbl}

\usepackage{tikz}
\usepackage{enumerate}

\usepackage{makecell}
\usepackage{subcaption}

\usepackage{algorithm}
\usepackage{algpseudocode}
\usepackage{enumitem}

\begin{document}

\title{
FireSentry: A Multi-Modal Spatio-temporal Benchmark Dataset for Fine-Grained Wildfire Spread Forecasting
}


\author{Nan Zhou}
\affiliation{%
  \institution{Shenzhen International Graduate School, Tsinghua University}
  \city{Shenzhen}
  \country{China}}
\email{zhoun24@mails.tsinghua.edu.cn}

\author{Huandong Wang}
\affiliation{%
  \institution{Department of Electronic Engineering, Tsinghua University}
  \city{Beijing}
  \country{China}}
\email{wanghuandong@tsinghua.edu.cn}

\author{Jiahao Li}
\affiliation{%
  \institution{Shenzhen International Graduate School, Tsinghua University}
  \city{Shenzhen}
  \country{China}}
\email{li-jh23@mails.tsinghua.edu.cn}

\author{Han Li}
\affiliation{%
  \institution{Shenzhen International Graduate School, Tsinghua University}
  \city{Shenzhen}
  \country{China}}
\email{h-li23@mails.tsinghua.edu.cn}


\author{Yali Song}
\affiliation{%
  \institution{College of Soil and Water Conservation, Southwest Forestry
University}
  \city{Yunnan}
  \country{China}}
\email{songyali@swfu.edu.cn}

\author{Qiuhua Wang}
\affiliation{%
  \institution{College of Civil Engineering, Southwest Forestry
University}
  \city{Yunnan}
  \country{China}}
\email{qhwang2010@swfu.edu.cn}

\author{Yong Li}
\affiliation{%
  \institution{Department of Electronic Engineering, Tsinghua University}
  \city{Beijing}
  \country{China}}
\email{wanghuandong@tsinghua.edu.cn}

\author{Xinlei Chen}
\affiliation{%
  \institution{Shenzhen International Graduate School, Tsinghua University}
  \city{Shenzhen}
  \country{China}}
\email{chen.xinlei@sz.tsinghua.edu.cn}


\renewcommand{\shortauthors}{Nan Zhou et al.}

\begin{abstract}

Fine-grained wildfire spread prediction is crucial for enhancing emergency response efficacy and decision-making precision. 
However, existing research predominantly focuses on coarse spatiotemporal scales and relies on low-resolution satellite data, capturing only macroscopic fire states while fundamentally constraining high-precision localized fire dynamics modeling capabilities.
To bridge this gap, we present FireSentry, a provincial-scale multi-modal wildfire dataset characterized by sub-meter spatial and sub-second temporal resolution.
Collected using synchronized UAV platforms, FireSentry provides visible and infrared video streams, in-situ environmental measurements, and manually validated fire masks.
Building on FireSentry, we establish a comprehensive benchmark encompassing physics-based, data-driven, and generative models, revealing the limitations of existing mask-only approaches. 
Our analysis proposes FiReDiff, a novel dual-modality paradigm that first predicts future video sequences in the infrared modality, and then precisely segments fire masks in the mask modality based on the generated dynamics.
FiReDiff achieves state-of-the-art performance, with video quality gains of 39.2\% in PSNR, 36.1\% in SSIM, 50.0\% in LPIPS, 29.4\% in FVD, and mask accuracy gains of 3.3\% in AUPRC, 59.1\% in F1 score, 42.9\% in IoU, and 62.5\% in MSE when applied to generative models.
The FireSentry benchmark dataset and FiReDiff paradigm collectively advance fine-grained wildfire forecasting and dynamic disaster simulation.
The processed benchmark dataset is publicly available at:
\href{https://github.com/Munan222/FireSentry-Benchmark-Dataset}{https://github.com/Munan222/FireSentry-Benchmark-Dataset}.

\end{abstract}



\maketitle

\section{Introduction}

\begin{figure*}[t!]
    \centering
    \begin{subfigure}[b]{0.33\textwidth}
        \centering
        \includegraphics[height=5cm]{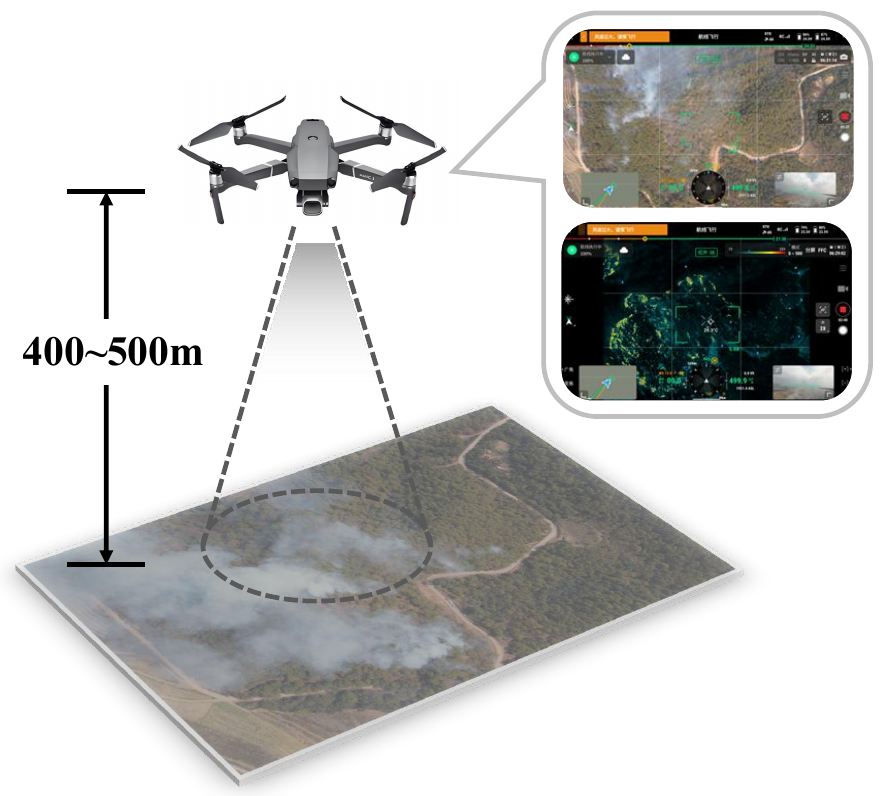}
        \caption{Fire dynamics monitoring via synchronized UAV-based visible and infrared spectra.}
        \label{fig:gptrack_arch_a}
    \end{subfigure}
    \hfill
    \begin{subfigure}[b]{0.65\textwidth}
        \centering
        \includegraphics[height=5cm]{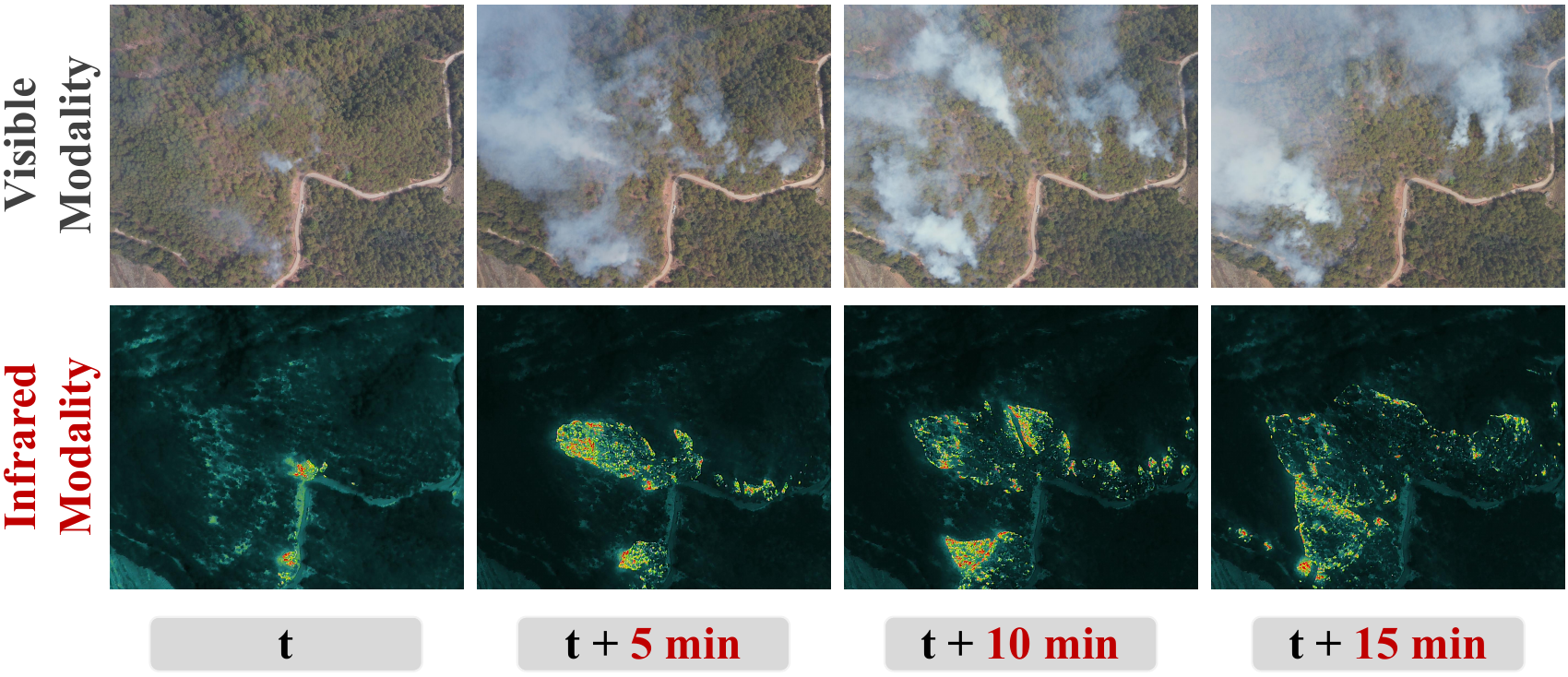}
        \caption{
        Fire spread visualization in our FireSentry dataset over a 200,000 square meters area in dual visible and infrared spectra at 0.5 meters spatial and 5‑minute temporal resolution.
        }
        \label{fig:gptrack_arch_b}
    \end{subfigure}
    \caption{
    Wildfire data acquisition and dual-spectral video visualization.
    }
    \label{fig:gptrack_architecture}
\end{figure*}

Accurate and real-time fine-grained wildfire spread prediction is crucial for facilitating efficient evacuations and optimizing emergency responses. However, the highly dynamic and complex nature of wildfires poses significant challenges to accurate forecasting.

To address this challenge, existing approaches fall into three categories: 
physics-based models simulate fire dynamics using principles like fluid mechanics and heat transfer, yet struggle with real-world complexity~\cite{mandel2011coupled}; 
data-driven models target large-scale temporal fire prediction and are heavily dependent on low-resolution satellite imagery, necessitating architectural adaptations for fine-grained wildfire prediction tasks~\cite{wang2022predrnn,gao2022earthformer};
and generative techniques (particularly world model-based video generation) show considerable promise but remain largely unexplored for wildfire contexts~\cite{Ding25WorldModels,ho2020denoising}.
Critically, wildfires exhibit flashover behavior, characterized by meter-scale expansions within minutes (Figure~\ref{fig:gptrack_architecture}), which necessitates high-resolution prediction for timely intervention.
However, existing datasets typically offer only hundred-meter spatial resolution and hourly temporal resolution (Table~\ref{tab:fire_datasets}), fundamentally lacking the spatio-temporal granularity required for such modeling.

To bridge this gap, we introduce FireSentry, a novel benchmark dataset designed specifically for fine-grained wildfire spread prediction. 
Covering five distinct regions within a single province, the dataset provides dual-modality visual data in both visible light and infrared spectra, along with environmental data.
To support model development, we generate fire masks from infrared videos using advanced semantic segmentation algorithms.
Mask quality is ensured through a human verification protocol: annotators create a validation subset by referencing infrared videos as the primary source and visible-light videos as auxiliary inputs (providing smoke dispersion and vegetation context). 
Quantitative evaluation via pixel-wise comparison between algorithmic masks and human-verified annotations yields an average accuracy of 0.925, mean Intersection-over-Union (mIoU) of 0.696, with commission and omission errors of 0.076 and 0.015 respectively.

Leveraging its high spatio-temporal resolution and multi-modal capabilities, FireSentry establishes a robust foundation for fine-grained wildfire propagation modeling. 
Building upon this dataset, we propose FiReDiff, a novel predictive paradigm. Diverging from conventional mask-level approaches ~\cite{li2024sim2real, huot2020deep}, FiReDiff innovatively employs a dual-stage architecture: first performing video prediction in the infrared modality, then executing fire mask segmentation within the mask modality.
This paradigm deeply integrates video generation with semantic understanding, achieving significant performance breakthroughs while pioneering new pathways for wildfire forecasting research.

We conduct a comprehensive benchmark evaluation on FireSentry, comparing physics-based, data-driven, and generative models, as well as our proposed FiReDiff paradigm. Experimental results demonstrate that FiReDiff consistently outperforms state-of-the-art~(SOTA) baselines across multiple evaluation metrics, highlighting its effectiveness and potential in fine-grained wildfire prediction tasks.
In summary, the contributions of this work are as follows:
\begin{itemize}[nosep,leftmargin=*]
    \item We present FireSentry, a multi-modal wildfire dataset that enables meter-level and minute-scale dynamics modeling. It integrates synchronized UAV-captured visible and infrared video streams, a spatio-temporally calibrated environmental telemetry, and manually validated fire segmentation masks. 
    \item 
    We introduce FiReDiff, a generative model–based prediction paradigm that jointly optimizes infrared video prediction and mask segmentation. 
    By integrating spatio–temporal features from complementary infrared and mask modalities, FiReDiff mitigates key constraints of mask-only approaches and enhances prediction robustness.
    \item 
    We establish comprehensive benchmarking protocols comparing physics-based, data-driven, generative approaches, and our novel FiReDiff paradigm. Extensive experiments quantitatively demonstrate FiReDiff's superior spatio-temporal prediction accuracy.
\end{itemize}

\begin{table*}[t!]
\centering
\renewcommand{\arraystretch}{1.45} 
\setlength{\tabcolsep}{0.9pt}
\caption{Comparison of fire dataset characteristics.}
\begin{tabular}{c|ccccccccc}
\Xhline{2\arrayrulewidth} 
\textbf{Dataset} & \textbf{Coverage} & \textbf{Task} & \textbf{\makecell{Spa. \\ Reso.}} & \textbf{\makecell{Temp. \\ Reos.}} & \textbf{\makecell{Areas\\(km\textsuperscript{2})}} & \textbf{Period} & \textbf{Fig/Video} & \textbf{Device} & \textbf{Mod.} \\ \Xhline{2\arrayrulewidth} 
Sim2Real-Fire~\cite{li2024sim2real} & Worldwide & \makecell{Fire Mask Forecast\\ Fire Mask Backtrack} & 30 m & 1 hour & 20,000,000 & 2013-2023 &  Fig & Satellite & 5 \\ \hline
FireSpreadTS~\cite{gerard2023wildfirespreadts} & USA &\makecell{Fire Mask Forecast\\ Fire Mask Backtrack} & 375 m & 1 day & 9,834,000 & 2018-2021 & Fig & Satellite & 4 \\ \hline
Mesogeos~\cite{kondylatos2023mesogeos} & Mediterranean & \makecell{Fire Mask Forecast\\ Danger Forecast} & 1000 m & 1 day & 9,000,999 & 2006-2022 & Fig & Satellite & 4 \\ \hline
FireDB~\cite{singla2021wildfiredb}& USA & Fire Intensity Forecast & 375 m & 1 day & 9,834,000 & 2012-2017 & Fig & Satellite & 4 \\ \hline
Next Day fire~\cite{huot2022next}& USA & Fire Mask Forecast & 1 km & 1 day & 9,834,000 & 2012-2020 &   Fig & Satellite & 4 \\ \hline
SeasFire Cube~\cite{karasante2025seasfire}& Worldwide & Fire Behavior & 500 m & 1 day & 149,000,000 & 2001-2017 &   Fig & Satellite & 4 \\ \hline
CFSDS~\cite{barber2024canadian}& Canada & Fire Behavior & 180 m & 1 day & 9,985,000 & 2002–2021 &   Fig & Satellite & 4 \\ \hline
BA-ONFIRE~\cite{gincheva2024monthly} & Worldwide & Fire Behavior & 111 km & 1 month & 149,000,000 & 1950-2021 & Fig & Satellite & 1 \\ \hline
GlobFire~\cite{artes2019global} & Worldwide & Fire Behavior & 500 m & 1 day & 149,000,000 & 2001-2017 & Fig & Satellite & 1 \\ \hline
Re-FWI~\cite{vitolo20191980} & Canada & Fire Behavior & 80 km & 1 day & 9,985,000 & 1980-2018 &   Fig & Satellite & 7 \\ \hline
GFBS~\cite{he2023global} & Worldwide & Fire Behavior & 30 m & 16 days & 149,000,000 & 2003-2016 & Fig & Satellite & 1 \\ \hline
FPA FOD-A~\cite{pourmohamad2023physical} & USA & Fire Behavior & 4 km & 1 day & 9,834,000 & 1992-2020 &   Fig & Satellite & 4 \\ \hline
GFED5~\cite{chen2023multi} & Worldwide & Fire Behavior & 500 m & 1 month & 149,000,000 & 1997-2020 & Fig & Satellite & 1 \\ \hline
CloCAB~\cite{hall2024glocab} & Worldwide & Fire Behavior & 27.7 km & 1 month & 149,000,000 & 2002-2020 & Fig & Satellite & 3 \\ \hline
PT-FireSprd~\cite{benali2023portuguese} & Portugal & \makecell{Fire Behavior\\ Fire Mask Forecast\\ Danger Forecast} & 4 km & 14 hours & 92,150 & 2015-2021 & Fig & Satellite & 1 \\ \hline
NSMC-H8~\cite{chen2023adapted} & China & Fire Monitoring & 3 km & 1 hour & 9,597,000 & 2019-2021 & Fig & Satellite & 1 \\ \hline
WCU-US~\cite{gray2018weekly} & Western USA & Danger Forecast & 250 m & 5 weeks & 4,200,000 & 2005-2017 & Fig & Satellite & 1 \\ \hline
ECUF~\cite{kim2024estimating} & Regions in South Korea & \makecell{Danger Forecast\\ Casualty Prediction} & 1 km & 1 month & 605 & 2017-2021 &   Fig & Satellite & 5 \\ \hline
LCLQ~\cite{pei2023using} & Regions in China & Danger Forecast & 500 m & 12 hour & 16,411 & 2015-2020 &   Fig & Satellite & 4 \\ \hline
CaBuAr~\cite{cambrin2023cabuar} & Regions in USA & Fire Mask Forecast & 20 m & 1 year & 450,000 & 2015-2022 & Fig & Satellite & 1 \\ \hline
GABAM~\cite{long201930} & Worldwide & Fire Behavior & 30 m & 1 year & 149,000,000 & 1990-2021 & Fig & Satellite & 1 \\ \hline
Fire Atlas~\cite{andela2019global} & Worldwide & Fire Behavior & 500 m & 1 day & 149,000,000 & 2003-2016 & Fig & Satellite & 1 \\ \hline
MODIS~\cite{nasa_modis_fire} & Worldwide & \makecell{Fire Mask Forecast\\ Danger Forecast} & 1 km & 1 day & 149,000,000 & 2000-2024 &   Fig & Satellite & 3 \\ \hline
VIIRS~\cite{nasa_viirs_fire} & Worldwide & \makecell{Fire Mask Forecast\\ Danger Forecast} & 375 m & 12 hours & 149,000,000 & 2012-2024 &   Fig & Satellite & 3 \\ \hline
NOAA HMS Fire~\cite{noaa_hms_fire_2024} & North America & Danger Forecast & 2 km & 1 day & 24,710,000 & 2003-2024 &   Fig & Satellite & 3 \\ \hline
NOAA HMS Smoke~\cite{noaa_hms_fire_2024} & North America & Danger Forecast & 2 km & 1 day & 24,710,000 & 2005-2024 & Fig & Satellite & 1 \\ \hline
GOES fire~\cite{noaa_goesr_2024} & Western Hemisphere & \makecell{Fire Mask Forecast\\ Danger Forecast} & 2 km & 5 mins & 61,000,000 & 2017-2024 &   Fig & Satellite & 4 \\ \hline
NIFC WP~\cite{nifc_wfigs_2024} & USA & Fire Mask Forecast & 2 km & 5 mins & 9,834,000 & 2000-2024 & Fig & Satellite & 1 \\ \hline
FRY~\cite{laurent2018fry} & Worldwide & Fire Behavior & 500 m & 1 day & 149,000,000 & 2005-2011 & Fig & Satellite & 1 \\ \hline
FireCanada~\cite{sayad2019predictive} & Canada & Fire Behavior & 1 km & 1 day & 61 & 2014.8 &   Fig & Satellite & 3 \\ \Xhline{2\arrayrulewidth}
\textbf{FireSentry} & Regions in China & \makecell{Fire Video Prediction\\ Infrared Video Prediction} & \textbf{0.5 m} & \textbf{1 s} & 0.7 & 2025.2 & \textbf{Video} & \textbf{Drone} & 4 \\ 
\Xhline{2\arrayrulewidth} 
\end{tabular}
\label{tab:fire_datasets}
\end{table*}

\begin{figure*}[t!]
    \centering
    \includegraphics[width=0.95\textwidth]{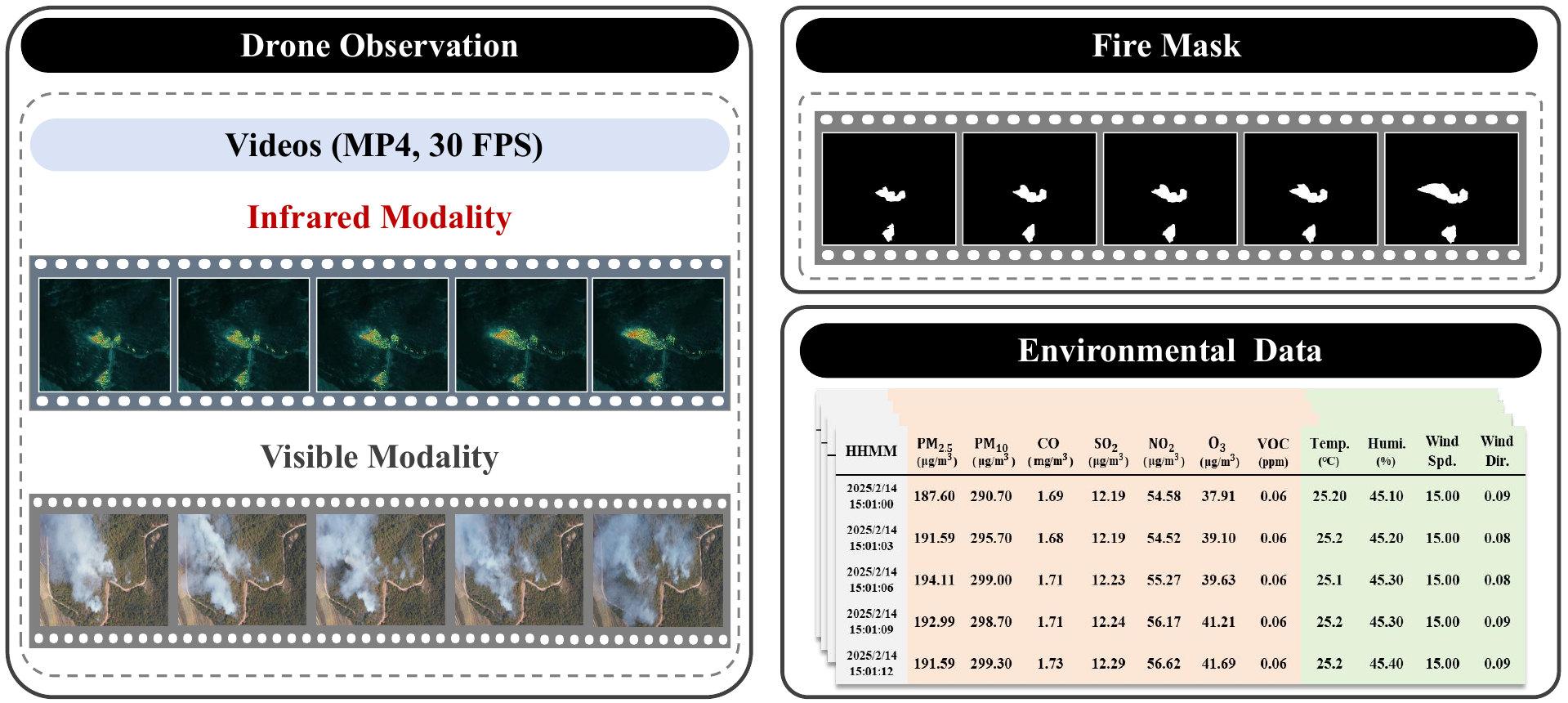}
    \caption{Dual-modal UAV-captured video streams, fire mask, and environmental data in the FireSentry dataset.}
    \label{fig:dataset}
\end{figure*}

\section{Related Work}

We systematically review the research landscape of wildfire spread prediction, 
and reveal a fundamental limitation in existing studies: the predominant focus on macro-scale spatiotemporal prediction, which hinders fine-grained fire spread modeling.

\subsection{Existing Efforts on Fire Task}

Existing research on fire spread prediction predominantly falls into three methodological categories. 
The first category encompasses physics-based methods, such as WRF-SFIRE~\cite{mandel2011coupled}, FARSITE~\cite{finney1998farsite}, and WFDS~\cite{meerpoel2023modeling}. These models simulate fire propagation based on principles of fluid dynamics, combustion, and heat transfer. 
While physically interpretable, these models often suffer from limited predictive accuracy and generalization in complex real-world scenarios.

The second category consists of data-driven models that 
learn spatio-temporal wildfire dynamics from historical patterns~\cite{wang2022predrnn,gao2022earthformer}. Among these, Sim2Real~\cite{li2024sim2real} employs Transformer architectures for sequence prediction; DL-fire~\cite{huot2020deep} implements segmentation via U-Net and LSTM hybrids; and WSTS+~\cite{lahrichi2025advancing} advances fire spread sequence modeling under single-day and multi-day input scenarios. 
However, these models primarily target large-scale fire prediction at temporal scales and exhibit heavy reliance on low-resolution satellite imagery for training. For fine-grained wildfire forecasting scenarios, existing frameworks require architectural modifications to accommodate high-precision demands.

The third category includes generative models,
particularly world model-based video generation techniques. These methods predict future states by learning historical dynamics~\cite{Ding25WorldModels,Wang25WISA}, with diffusion models emerging as a pivotal branch~\cite{ho2020denoising} that synthesizes visual content through iterative denoising. Notable implementations include MCVD's conditional video generation framework~\cite{voleti2022mcvd} and VDT's enhanced architecture with spatio-temporal attention~\cite{lu2023vdt}. 
Despite their demonstrated potential in dynamic scene generation, these models remain unexplored for wildfire spread prediction. 
Our work pioneers the integration of diffusion-based world modeling into fire forecasting, targeting high-resolution spatio-temporal modeling of fire behavior dynamics.

\subsection{Limitation of Existing Dataset}


Wildfire propagation exhibits pronounced flashover dynamics, enabling fires to expand rapidly across hundreds of thousands of square meters within minutes. 
Our drone-captured case documents fire evolution across 200,000 m² within a 15-minute interval (Figure~\ref{fig:gptrack_architecture}), while resolution limitations in current datasets constrain capturing such sub-minute granular dynamics.


As summarized in Table~\ref{tab:fire_datasets}, current wildfire datasets primarily support global/national-scale monitoring tasks, with heavy reliance on remote sensing satellites. These datasets exhibit two fundamental constraints: spatial resolutions typically at kilometer-scale 
and temporal resolutions at daily frequency.
Crucially, they deliver only single-modality fire masks without multi-modal infrared and visible-light observations, which is essential for capturing fire intensity dynamics, smoke dispersion patterns, and vegetation changes.


These limitations directly impair wildfire modeling since coarse-grained data cannot support short-term fire evolution analysis and modality deficiency hinders deep diagnostics.
In real emergency scenarios, such datasets severely compromise prediction model responsiveness and accuracy, failing to support precise decision-making for firefighting operations. 
Establishing high spatio-temporal resolution multi-modal datasets has thus become imperative to both enhance disaster response efficacy and safeguard lives while mitigating ecological damage.

\begin{figure*}[t!]
    \centering
    \begin{subfigure}[b]{0.3\textwidth}
        \centering
        \includegraphics[height=5.5cm]{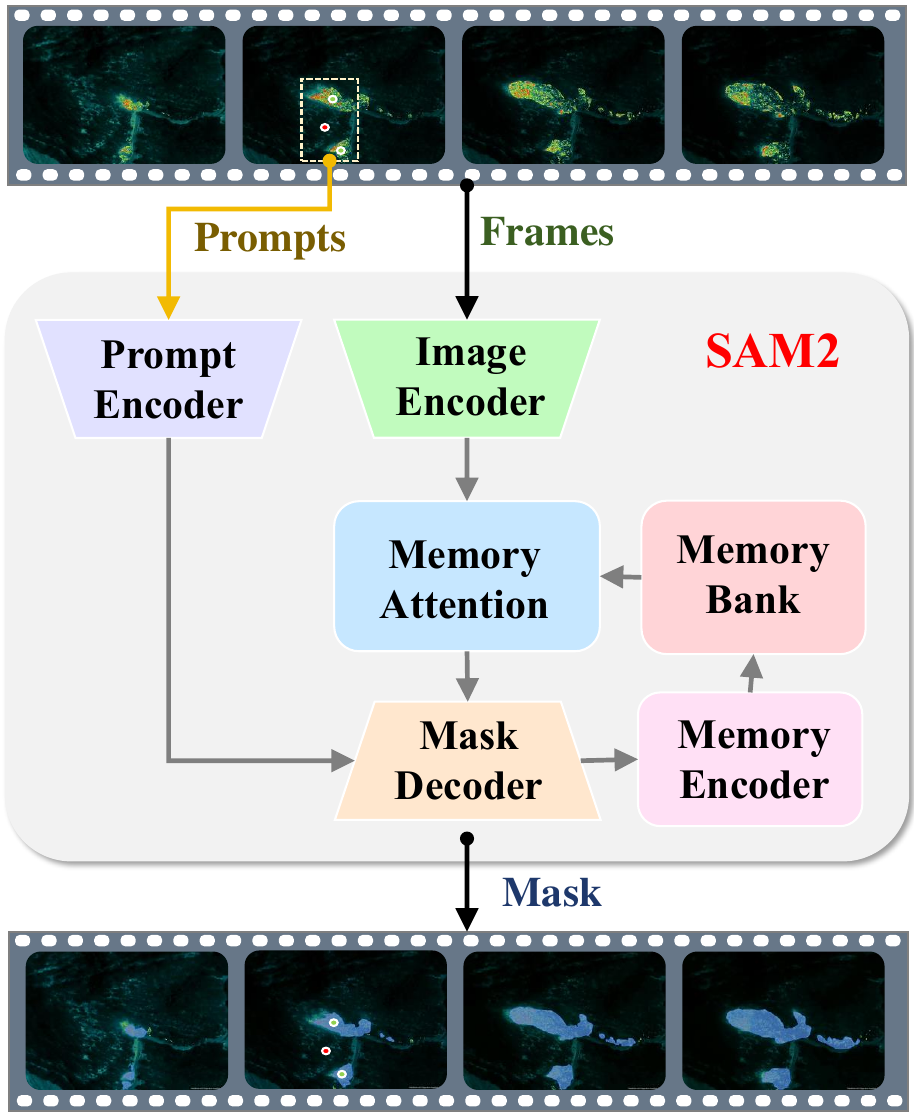}
        \caption{Algorithmic annotation: SAM2 generates fire masks for infrared videos.}
        \label{fig:sam2}
    \end{subfigure}
    \hfill
    \begin{subfigure}[b]{0.68\textwidth}
        \centering
        \includegraphics[height=5.5cm]{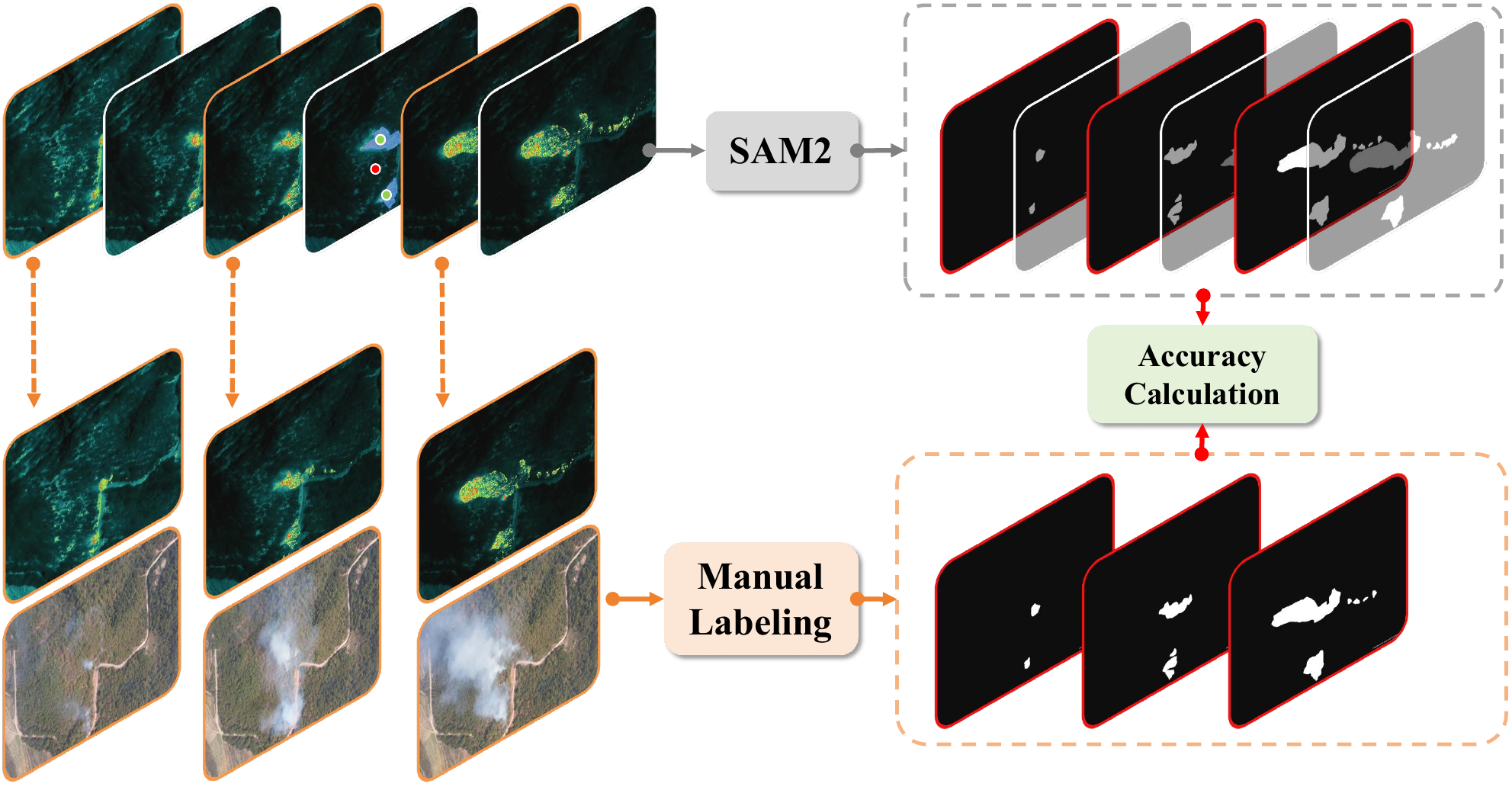}
        \caption{Manual verification: human annotation of selected infrared frames assisted by visible-spectrum data with algorithmic accuracy evaluation.}
        \label{fig:accuracy}
    \end{subfigure}
    \caption{Mask annotation pipeline.}
    \label{fig:mask pipeline}
\end{figure*}

\begin{figure*}[t!]
    \centering
    \begin{subfigure}[b]{0.2\textwidth}
        \centering
        \includegraphics[height=2.75cm]{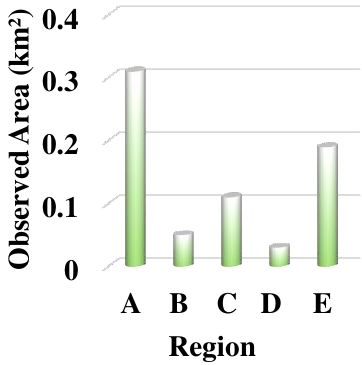}
        \caption{Land coverage area.}
        \label{fig:cover}
    \end{subfigure}
    \begin{subfigure}[b]{0.79\textwidth}
        \centering
        \includegraphics[height=2.75cm]{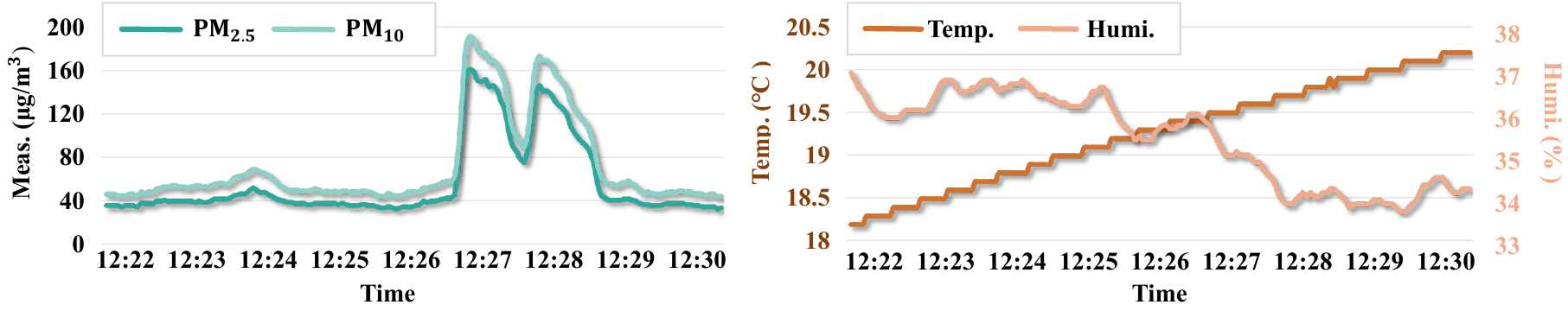}
        \caption{Temporal variation of environmental data in region A.}
        \label{fig:env var}
    \end{subfigure}
    \caption{FireSentry dataset analysis. 
    }
    \label{fig:data stat}
\end{figure*}

\section{FireSentry Dataset}

\subsection{Data Description}

As shown in Figure~\ref{fig:dataset}, the FireSentry dataset provides synchronized multi-modal video streams comprising visible-light and infrared modalities, alongside corresponding fire segmentation masks and environmental monitoring metrics.
Critically, all modalities have been rigorously co-registered in both space and time, ensuring spatial consistency and accurate cross-modal alignment.
The dataset covers five representative high-risk fire regions (A–E) within a provincial-level administrative area, collected during the critical 2025 fire-prevention season.

\paragraph{Drone Observation.}
As depicted in Figure~\ref{fig:gptrack_arch_a}, drones maintain static hover positions at 400–500 meters above ground level to capture wildfire propagation dynamics, continuously recording synchronized infrared and visible-spectrum video streams.
Our dataset features structured tri-phase observations per monitored wildfire zone: pre-event scanning, prolonged event monitoring (1–2 hours duration), and post-event scanning. Raw video streams are segmented into temporally contiguous clips of fixed 16-minute durations. Per-zone observations yield 6–23 temporally continuous video segments, with quantity scaling proportionally to monitoring duration. All segments maintain consistent $1280 \times 1024$ resolution at 30 frames per second (FPS), resulting in 1 second temporal resolution and 0.5 meter/pixel ground sampling distance.

\paragraph{Fire Mask.}
Visible-spectrum videos are susceptible to smoke interference and vegetation occlusion, compromising fire boundary delineation accuracy.
Conversely, infrared video streams effectively mitigate such visual obstructions, enabling unobstructed wildfire dynamics observation (Figure~\ref{fig:gptrack_arch_b}). 
To support model development, we generate fire masks from infrared streams using a SOTA semantic segmentation algorithm~\cite{ravi2024sam}.
For annotation quality assurance, we implement a dual-stage verification protocol: 
(i) Multiple annotators establish reference mask subsets using infrared frames as primary input, supplemented by visible-spectrum videos for smoke dispersion and vegetation coverage reference;
(ii) Geometric correspondence to human-annotated benchmarks is quantified through four pixel-level consistency metrics: accuracy, mIoU, commission error, and omission error.
The full annotation workflow is detailed in Section \ref{mask annot}.

\paragraph{Multi-source Environmental Characterization.}
Temporally, high-frequency environmental parameters are collected via sensor networks, while spatially, land vegetation parameters are retrieved from dedicated web platforms.

Environmental data, acquired via fixed sensors, constitute continuous time-series recordings capturing timestamps and eleven parameters: PM$_{2.5}$, PM$_{10}$, CO, SO$_2$, NO$_2$, O$_3$, VOC, temperature, humidity, wind speed, and wind direction. 
A dynamic sampling mechanism (adjustable between 3-30 seconds) is employed to enhance temporal feature diversity. The 3-second high-frequency mode precisely captures transient features during fire events, while the 30-second routine sampling establishes the environmental baseline. This design comprehensively spans time scales from second-level transients to minute-level pollution dispersion processes.

Land vegetation data are derived from \cite{gscloud_land_cover} and encompass five core parameters: 
land types including tree forest, rural grassland, pit pond, dry land, facility agricultural land, and other grassland; 
tree species composition categorized as Pinus kesiya, Quercus, other broad-leaved species, or non-tree; 
stem density (0–9,000 stems/ha) capturing vegetation gradients from sparse to dense stands; 
diameter at breast height (DBH) (0–24 meter) spanning seedling to over-mature growth stages; 
and canopy cover ratio (0–0.9) representing continuum from bare soil to closed-canopy conditions.

Spatial heterogeneity manifests in three dimensions:
(i) horizontal gradients in land cover, species composition, and stem density reveal ecosystem distribution patterns;
(ii) Vertical stratification quantified through canopy cover gradients and DBH distributions;
(iii) Biomass representation via DBH-stem density integration.
This multidimensional characterization provides critical inputs for wildfire risk modeling.

\begin{figure*}[t!]
    \centering
    \includegraphics[width=1 \textwidth]{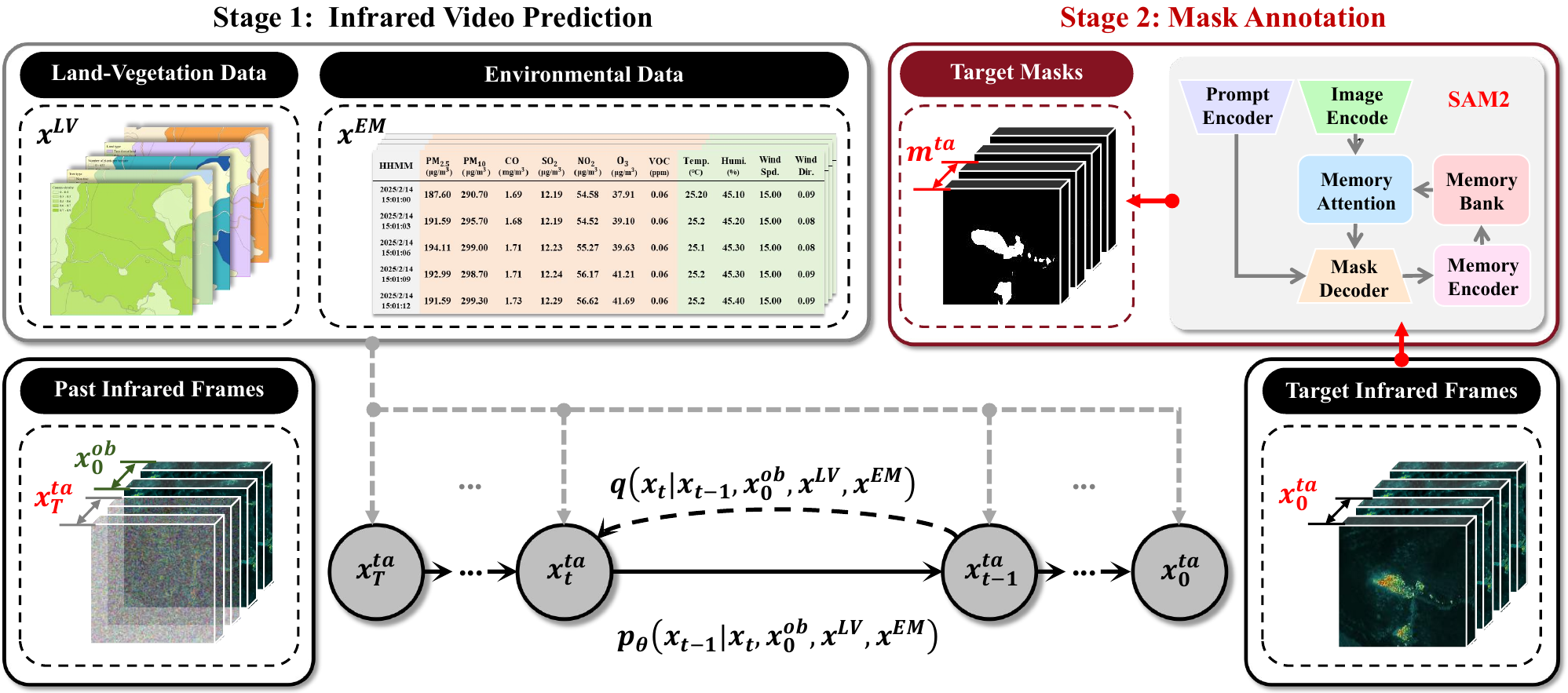}
    \caption{The framework of FiReDiff.}
    \label{fig:model}
\end{figure*}

\subsection{Mask Annotation Pipeline}
\label{mask annot}
The mask annotation pipeline for our dataset employs a dual-stage framework: algorithmic annotation and manual verification, as illustrated in Figure~\ref{fig:mask pipeline}. During the algorithmic annotation phase, we formulate fire region labeling in infrared videos as a promptable visual segmentation task. Leveraging the advanced SAM2 model~\cite{ravi2024sam}, multiple point prompts are provided on arbitrary video frames to designate target fire regions, enabling spatio-temporal mask prediction across entire video sequences (Figure~\ref{fig:sam2}), with model architecture detailed in Appendix~\ref{sec:appendix_sam}.

The manual verification phase constructs validation subsets using infrared imagery as primary source and visible-spectrum data as auxiliary input (providing contextual smoke dispersion and vegetation status), as specified in Appendix~\ref{sec:appendix_A1}. Quantitative comparison between manually annotated masks and SAM2-generated outputs follows the workflow in Figure~\ref{fig:accuracy}. 
Table~\ref{tab:accuracy_metrics} demonstrates SAM2's segmentation reliability, with 0.925 accuracy, 0.696 mIoU, 0.076 commission error, and 0.015 omission error.
Metric definitions and computations are formalized in Appendix~\ref{sec:appendix_A2}.

\subsection{Data Statistics}

The FireSentry dataset captures fire spread dynamics across five prescribed burn regions, with a total monitored area of 0.7 km² (spatial distribution shown in Figure~\ref{fig:cover}). Using region A as a representative case, Figure~\ref{fig:env var} demonstrates the temporal evolution of environmental parameters during the initial 8-minute fire propagation phase: At 12:22 during incipient fire front formation, escalating heat release rate drove progressive temperature increase ($\Delta$T = +2°C over 8 min) while combustion-induced moisture dissipation caused continuous humidity decline ($\Delta$RH = -3.1\% over 8 min). 

When the fire front reached the sensor location at 12:26, high-temperature ignition of volatile organic compounds emitted by vegetation triggered incomplete combustion, generating substantial inhalable particulate matter that drove concentrations to PM$_{\text{2.5}}^{\text{peak}} >$ 160 $\mu$g/m$^3$ and PM$_{\text{10}}^{\text{peak}} >$ 190 $\mu$g/m$^3$.
Following 120 seconds of sustained combustion, fuel depletion at the sensor site by 12:28 reduced combustion intensity, allowing atmospheric dispersion and sedimentation mechanisms to decrease particulate concentrations by 74.5\% (PM$_{2.5}$) and 71.2\% (PM$_{10}$) from peak values, while humidity stabilized with weakened thermal convection. This progression confirms particulate matter concentrations are thermodynamically governed by fire intensity, exhibiting positive correlation with temperature and negative correlation with humidity.

\begin{table*}[t!]
\centering
\caption{Model performance comparison on fire video prediction task in mask modality. 
Percentage improvements indicate the relative performance gains over generative models after adopting FiReDiff.
Bold denotes highest improvement.
}
\renewcommand{\arraystretch}{1.35}
\setlength{\tabcolsep}{0.5pt}
\begin{adjustbox}{width=\textwidth}
\begin{tabular}{c|c|cccc|cccc}
\Xhline{2\arrayrulewidth}
\multicolumn{2}{c|}{} & \multicolumn{4}{c|}{\textbf{Video Quality Metrics}} & \multicolumn{4}{c}{\textbf{Fire Mask Metrics}} \\ 
\cmidrule(lr){3-6} \cmidrule(lr){7-10}
\multicolumn{2}{c|}{\multirow{-2}{*}{\centering\textbf{Method}}} & \textbf{PSNR↑} & \textbf{SSIM↑} & \textbf{LPIPS↓} & \textbf{FVD↓} & \textbf{AUPRC↑} & \textbf{F1 score↑} & \textbf{IoU↑} & \textbf{MSE↓} \\ 
\Xhline{2\arrayrulewidth}
\makecell{Physical \\ Model} & WRF-SFIRE~\cite{mandel2011coupled} & 22.614 & 0.622 & 0.132 & 287.330 & 0.816 & 0.760 & 0.616 & 0.019 \\ 
\hline
\multirow{2}{*}{\centering\makecell{Data-driven \\ Model}} & Earthformer~\cite{gao2022earthformer} & 3.926 & 0.348 & 0.480 & 866.512 & 0.763 & 0.543 & 0.376 & 0.419 \\ 
& PredRNN~\cite{wang2022predrnn} & 9.674 & 0.754 & 0.183 & 724.213 & 0.643 & 0.789 & 0.807 & 0.118 \\ 
\hline
\multirow{4}{*}{\centering\makecell{Generative \\ Model}} & MCVD~\cite{voleti2022mcvd} & 24.444 & 0.637 & 0.026 & 112.509 & 0.956 & 0.897 & 0.863 & 0.040 \\ 
& STDiff~\cite{ye2024stdiff} & 17.705 & 0.468 & 0.042 & 355.57 & 0.941 & 0.821 & 0.705 & 0.151 \\ 
& VDT~\cite{lu2023vdt} & 23.268 & 0.856 & 0.017 & 44.587 & 0.984 & 0.823 & 0.477 & 0.008 \\ 
& DynamiCrafter~\cite{xing2024dynamicrafter} & 7.415 & 0.729 & 0.252 & 1451.692 & 0.946 & 0.259 & 0.483 & 0.230 \\ \hline
\multirow{4}{*}{\textbf{\centering\makecell{FiReDiff \\ Paradigm}}} 
& \textbf{MCVD*} 
& 27.514 (+12.6\%) & 0.867 (\textbf{+36.1\%}) & 0.019 (+34.6\%) & 101.644 (+9.7\%) 
& 0.988 (\textbf{+3.3\%}) & 0.907 (+1.1\%) & 0.901 (+4.4\%) & 0.031 (+22.5\%) \\ 
& \textbf{STDiff*} 
& 19.235 (+8.6\%)  & 0.546 (+16.7\%) & 0.021 (\textbf{+50.0\%}) & 298.55 (+16.0\%)  
& 0.962 (+2.2\%) & 0.843 (+2.7\%) & 0.810 (+14.9\%) & 0.143 (+5.3\%) \\ 
& \textbf{VDT*} 
& 28.841 (+24.0\%) & 0.985 (+15.1\%) & 0.009 (+47.1\%) & 38.857 (+12.9\%)  
& 0.996 (+1.2\%) & 0.874 (+6.2\%) & 0.660 (\textbf{+42.9\%}) & 0.003 (\textbf{+62.5\%}) \\ 
& \textbf{DynamiCrafter*} 
& 10.324 (\textbf{+39.2\%}) & 0.815 (+11.8\%) & 0.198 (+21.4\%) & 1024.364 (\textbf{+29.4\%})  
& 0.974 (+3.0\%) & 0.412 (\textbf{+59.1\%}) & 0.451 (+6.6\%) & 0.154 (+33.0\%) \\ 
\Xhline{2\arrayrulewidth}
\end{tabular}
\end{adjustbox}
\label{tab:mask pred}
\end{table*}

\subsection{Data Limitations}

This work represents a multi‑modal wildfire dataset captured from a UAV platform, markedly advancing fine‑grained fire evolution modeling by delivering high spatio-temporal resolution (0.5 meter at 1 second). 
While the current dataset encompasses a single‑month observation window across five prescribed‑burn regions and captures multi-dimensional environmental dynamics, it remains constrained in its ability to characterize cross‑seasonal fire behaviors.

To develop a more comprehensive and generalizable wildfire analysis framework, we propose a systematic expansion strategy for future work. Data collection will be extended to cover cross-quarter observations during 2026 to 2027, capturing seasonal climate variability drivers in both dry and rainy periods. 
Additionally, natural wildfire events from three representative ecoregions, including temperate forests, subtropical shrublands, and boreal forests, will be integrated to assess and improve the geographic transferability of our modeling approach.
These extensions will enable deeper insights into climate-driven fire dynamics and broader applicability across diverse ecosystems.


\section{A Generative Diffusion Paradigm for Wildfire Spread Prediction}

This section introduces a novel paradigm—FiReDiff, a generative diffusion world framework for wildfire prediction. The methodology first extracts rich spatio-temporal features from infrared video streams, then performs mask segmentation. 
Unlike conventional approaches focused solely on mask prediction, FiReDiff achieves breakthrough performance through its dual-stage feature learning mechanism. The framework is developed and validated on our custom-built FireSentry dataset.

As illustrated in Figure~\ref{fig:model}, FiReDiff consists of two primary stages: infrared video prediction (stage 1) and mask annotation (stage 2). 
In the following subsections, 
we delineate the foundational architecture of the diffusion framework, the conditional modeling mechanism for infrared video prediction, and its novel operational paradigm in wildfire spread forecasting.

\paragraph{Diffusion Model.}
We adopt the denoising diffusion probabilistic model (DDPM) framework for sample generation~\cite{ho2020denoising}.
A clean data sample $x_0 \sim p_{\text{data}}$ is progressively perturbed over $T$ steps via a forward diffusion process:
\begin{equation}
    q(x_t \mid x_{t-1}) = \mathcal{N}(x_t; \sqrt{1 - \beta_t} x_{t-1}, \beta_t I),
\end{equation}
with a closed-form marginal
$q(x_t \mid x_0) = \mathcal{N}(x_t; \sqrt{\bar{\alpha}_t} x_0, (1 - \bar{\alpha}_t) I)$, where $\bar{\alpha}_t = \prod_{s=1}^t (1 - \beta_s).$
The reverse process starts from Gaussian noise $x_T \sim \mathcal{N}(0, I)$ and generates data by denoising through a neural network $\boldsymbol{\epsilon}_\theta(x_t, t)$ trained with the following objective:
\begin{equation}
    L(\theta) = \mathbb{E}_{t, x_0, \boldsymbol{\epsilon}} 
    \left[ \left\| \boldsymbol{\epsilon} - \boldsymbol{\epsilon}_\theta\left( \sqrt{\bar{\alpha}_t} x_0 + \sqrt{1 - \bar{\alpha}_t} \boldsymbol{\epsilon} \,\middle|\, t \right) \right\|_2^2 \right].
\end{equation}

This noise estimation corresponds to a scaled score function $ \nabla_{x_t} \log q(x_t \mid x_0) = -\frac{1}{\sqrt{1 - \bar{\alpha}_t}} \boldsymbol{\epsilon}$,
which enables noise-conditional score-based data generation. The detailed derivation is provided in the Appendix \ref{sec:ddpm}.
Extending this distributional formulation to temporally aligned frames enables natural adaptation to unconditional video generation tasks. However, systematic approaches for conditional tasks such as video prediction remain under-explored.

\paragraph{Video Prediction via Conditional Diffusion.}
The framework comprises two stages: infrared video prediction and fire mask annotation. The former explicitly models the conditional distribution of future infrared frames, while the latter extracts segmentation masks from the generated infrared frames.

Suppose we have $L$ past frames, denoted as $x^\mathrm{ob}_0\in \mathcal{R}^{L\times X\times Y}$, where $X$ and $Y$ represent the spatial resolution of the images. 
The land-vegetation data only considers the spatial dimension, represented as $ x^\mathrm{LV}\in\mathcal{R}^{ X\times Y}$, and the environmental data only considers the time dimension, represented as $ x^\mathrm{EM} \in \mathcal{R}^{L}$. The goal is to predict the next $L'$ frames, denoted as $x_0^\mathrm{ta}\in\mathcal{R}^{L'\times X\times Y}$.

\textit{Infrared Video Prediction Stage.} We fuse $x^\mathrm{ob}_0$, $x^\mathrm{LV}$, and $x^\mathrm{EM}$ as multi-modal conditional inputs to dynamically predict $x_0^\mathrm{ta}$:
\begin{align}
    & L_{\mathrm{FiReDiff}}(\theta)
    =  \ \mathbb{E}_{t,[x^\mathrm{ob}_0,x_0^\mathrm{ta}]\sim p_{\mathrm{data}}, \boldsymbol{\epsilon}\sim\mathcal{N}(0,I)} \Big[
    \lVert \boldsymbol{\epsilon} \nonumber - \\
    & \boldsymbol{\epsilon}_\theta(
    \sqrt{\bar{\alpha}_t}x_0^\mathrm{ta}
    + \sqrt{1 - \bar{\alpha}_t}\boldsymbol{\epsilon}
    \mid x^\mathrm{ob}_0,x^\mathrm{LV},x^\mathrm{EM},t)
    \rVert^2 
    \Big].
    \label{eq:6}
\end{align}

Our trained model employs an efficient block-wise auto-regressive strategy to generate high-fidelity frames, deeply parsing spatio-temporal correlations between environmental conditions and historical predictions while integrating multi-source contextual features. This integrated approach mines latent fire-spread patterns from heterogeneous environmental inputs through synergistic feature extraction.

\textit{Mask Annotation Stage.}
FiReDiff leverages the SAM2 module to transform generated infrared prediction frames $x_0^{ta}$ into precise fire masks 
$\hat{m}^{ta}$ , conditioned on input multi-point prompts:
\begin{equation}
    \hat{m}^{ta} = \mathrm{SAM2}\big(x_0^{ta}, \text{Prompts}\big).
\end{equation}

These two stages form a novel cascade that converts raw environmental data and historical infrared inputs into future infrared predictions, which are then transformed into fire segmentation masks, effectively enhancing wildfire situational awareness.
Implementation details appear in Algorithm \ref{algo 1} and \ref{algo 2} of Appendix \ref{appendix_algo}.

\section{Experimental Results}

\begin{figure}[t!]
    \centering
    \includegraphics[width=0.47\textwidth]{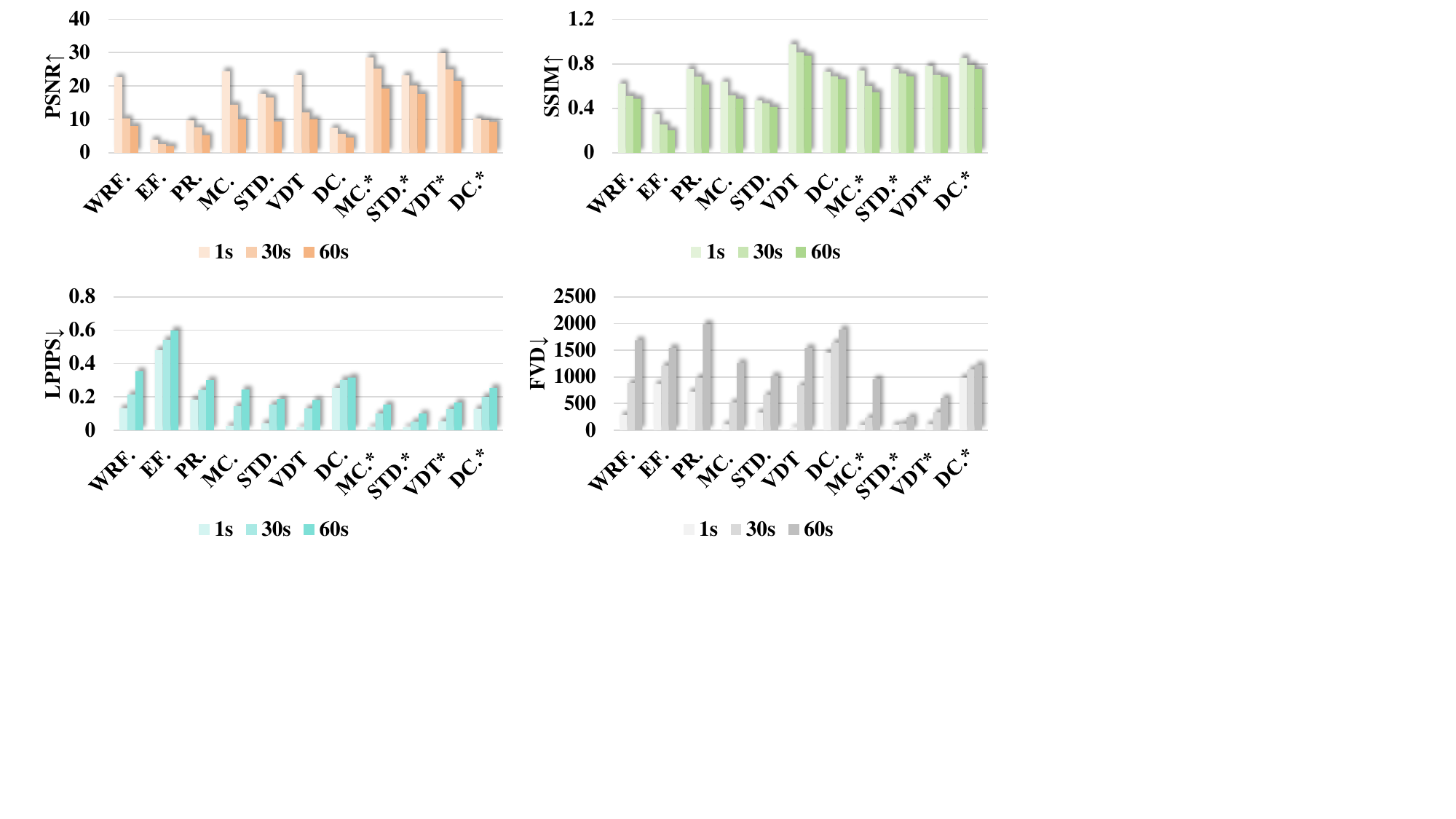}
    \caption{Comparative model performance across temporal resolutions. Abbreviations: WRF. = WRF-SFIRE,EF. = Earthformer, PR. = PredRNN, MC. = MCVD, STD. = STDiff, DC. = DynamiCrafter. * indicates generative models that implement the FiReDiff paradigm.}
    \label{fig:time}
\end{figure}

\subsection{Experimental Setup}

\paragraph{Experimental Task.}
Departing from conventional mask-based methods, FiReDiff pioneers a dual-modality framework that leverages both infrared and mask modalities for fire mask prediction, marking a modality-level paradigm shift.
For prediction tasks, each input contains 10 consecutive historical frames, with the objective of forecasting the subsequent 10 frames. 
Detailed experimental configurations are provided in Appendix~\ref{resolution}.

\paragraph{Benchmark Models.}
We conduct comprehensive benchmarking on FireSentry across four model categories: physical, data-driven, generative, and the FiReDiff paradigm. 
For the first three categories, we adapt seven representative open-source SOTA models (Table~\ref{tab:mask pred}) through input channel alignment, marking the first effort to repurpose image segmentation and video generation models for video-based fire forecasting.
Implementation details are provided in Appendix~\ref{app_benchmark}. For the fourth category, we implement FiReDiff paradigm's dual-phase strategy within the generative model: infrared video prediction followed by fire mask segmentation from generated frames.

\paragraph{Evaluation Metrics}
For the fire video prediction task, we establish a dual-modality evaluation framework that jointly assesses segmentation accuracy and video quality. Segmentation quality is quantified through four metrics: AUPRC, F1 score, IoU, and MSE; video quality is evaluated via PSNR, SSIM, LPIPS, and FVD. 
This unified setup supports comprehensive assessment of both frames and sequences (see Appendix~\ref{app_meric} for details).

\subsection{Overall Performance}


As shown in Table~\ref{tab:mask pred}, the FiReDiff paradigm substantially outperforms SOTA methods, achieving superior results in both video quality and fire mask segmentation accuracy. Moreover, FiReDiff notably boosts the overall performance of generative models across modalities.
Qualitative visualizations of predicted mask sequences are presented in Figure~\ref{fig:visual_mask} (Appendix~\ref{vis_mask}), while quantitative results and sample generations for the infrared modality are provided in Table~\ref{tab:infra pred} and Figure~\ref{fig:visual_infra} (Appendix~\ref{vis_infra}). These experimental findings consistently demonstrate FiReDiff’s strong performance across both infrared and mask modalities.

Additionally, we evaluate fire spot/line recognition accuracy under permissible geolocation errors of 1.5-3 meters. Results demonstrate average accuracy rates of 92.8\% for fire spots and 87.9\% for fire lines, with detailed analysis provided in Appendix~\ref{fire_line_point}.

\begin{table}[t!]
\centering
\caption{Performance comparison under environmental meteorological (EM) and land‑vegetation (LV) modality ablations. Bold indicates the best, and underline indicates the second-best. Percentages in parentheses denote the performance drop relative to the full-modality setting.}
\renewcommand{\arraystretch}{1.35}
\setlength{\tabcolsep}{3pt}
\fontsize{9}{11}\selectfont
\begin{tabular}{c|ccc|ccc}
\Xhline{2\arrayrulewidth}
\multirow{2}{*}{\textbf{Method}} & \multicolumn{3}{c|}{\textbf{PSNR}~$\uparrow$} & \multicolumn{3}{c}{\textbf{MSE}~$\downarrow$} \\
\cmidrule(lr){2-4} \cmidrule(lr){5-7}
& w/o LV & w/o EM & Full & w/o LV & w/o EM & Full \\
\Xhline{2\arrayrulewidth}
\textbf{EF} & 
\makecell{3.524\\(-10.2\%)} & 
\makecell{3.124\\(-20.4\%)} & 
3.926 & 
\makecell{0.468\\(-11.7\%)} & 
\makecell{0.598\\(-42.7\%)} & 
0.419 \\
\hline
\textbf{PR} & 
\makecell{9.062\\(-6.3\%)} & 
\makecell{8.432\\(-12.8\%)} & 
9.674 & 
\makecell{0.195\\(-65.3\%)} & 
\makecell{0.264\\(-123.7\%)} & 
0.118 \\
\hline
\textbf{STD} & 
\makecell{\underline{16.854}\\(-4.8\%)} & 
\makecell{\underline{16.125}\\(-8.9\%)} & 
\underline{17.705} & 
\makecell{\underline{0.168}\\(-12.0\%)} & 
\makecell{\underline{0.187}\\(-24.7\%)} & 
\underline{0.150} \\
\hline
\textbf{MC} & 
\makecell{\textbf{23.034}\\(-5.8\%)} & 
\makecell{\textbf{22.126}\\(-9.5\%)} & 
\textbf{24.444} & 
\makecell{\textbf{0.087}\\(-11.75\%)} & 
\makecell{\textbf{0.125}\\(-212.5\%)} & 
\textbf{0.040} \\
\Xhline{2\arrayrulewidth}
\end{tabular}
\label{tab:modality_ablation}
\end{table}

\subsection{Spatio-temporal Resolution Scaling Analysis}


Models are trained at 1-second inter-frame physical time intervals and evaluated under extrapolation tasks at 1 second, 30 seconds, and 60 seconds intervals for equivalent frame lengths. 
As shown in Figure~\ref{fig:time}, performance exhibits systematic degradation with increasing intervals, primarily due to amplified prediction state uncertainty at longer timescales.
Critically, FiReDiff consistently outperforms SOTA baselines while demonstrating enhanced robustness across multi-temporal-scale predictions.

To further validate the spatial stability of the FiReDiff paradigm, we compared DynamiCrafter with DynamiCrafter* (which applies FiReDiff) across varying spatial resolutions. Results demonstrate that even as resolution doubles, FiReDiff consistently enhances model robustness by an average of 45.4\%; detailed results are provided in Appendix~\ref{spatial_exp}.

\subsection{Multi-modal Data Integration Strategies}

To quantify the importance of environmental modalities, we systematically analyze performance impacts from removing environmental meteorological (EM) or land-vegetation (LV) data. Validated through four architecturally diverse cross-modal fusion models (MCVD, STDiff, Earthformer, PredRNN), Table~\ref{tab:modality_ablation} reveals consistent degradation patterns: LV removal causes average PSNR reductions of 21.0\% and MSE increases of 25.19\%, while EM removal induces 12.9\% PSNR decline and 100.90\% MSE surge. These results conclusively demonstrate both modalities' irreplaceable contributions to fire dynamics modeling.

\section{Conclusion}
We present FireSentry, a UAV-collected multi-modal dataset for fine-grained wildfire spread prediction, and propose FiReDiff, an innovative cross-modal paradigm that first generates future infrared sequences and then automatically segments fire masks.
Experimental validation shows that FiReDiff outperforms physical, data-driven, and generative baselines, achieving notable performance gains of 39.2\% in PSNR, 36.1\% in SSIM, 50.0\% in LPIPS, 29.4\% in FVD, 3.3\% in AUPRC, 59.1\% in F1 score, 42.9\% in IoU, and 62.5\% in MSE, when applied to generative models.
Spatio-temporal resolution scaling experiments verify FiReDiff's robustness across diverse granularities, while multi-modal ablations highlight the indispensable role of environmental meteorological telemetry and land-vegetation indices. 
Collectively, our results establish a novel video-driven paradigm for wildfire forecasting, signifying a notable advancement in fine-grained fire modeling.

\newpage

\bibliographystyle{ACM-Reference-Format}
\bibliography{refer}

\clearpage
\appendix

\section{Additional Information About the Mask Annotation Pipeline}

\begin{figure*}[h]
    \centering
    \includegraphics[width=0.9\textwidth]{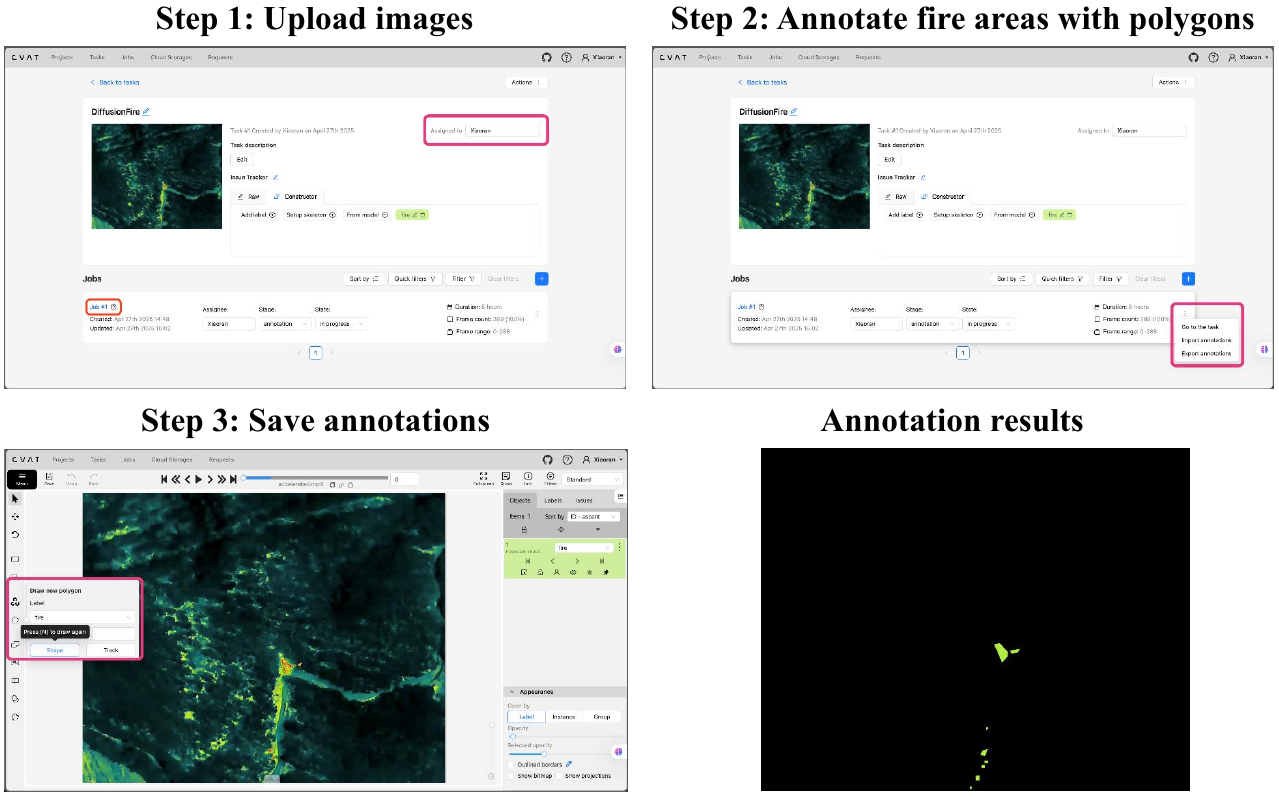}
    \caption{Fire mask annotation workflow using Computer Vision Annotation Tool.}
    \label{fig:a1}
\end{figure*}

\subsection{Manual Inspection in the Data Annotation and Quality Control Process}
\label{sec:appendix_sam}

As illustrated in Figure~\ref{fig:sam2}, the overall annotation workflow of SAM2 operates as follows: for a given frame, segmentation prediction is conditioned on the current prompt and previously observed memory. The video is processed in a streaming manner, where prompts are fed into the Prompt Encoder, and frames are sequentially passed through the Image Encoder. These frame features are then associated with the memory of target objects from previous frames via Memory Attention. The Mask Decoder optionally accepts prompt inputs and generates the segmentation mask for the current frame. Finally, the Memory Encoder fuses the prediction results with features from the Image Encoder (not shown in the figure), converting them into memory representations that are stored in the Memory Bank for future reference.
Here is a brief overview of the six core modules of SAM2:

\begin{itemize}[nosep,leftmargin=*]
    \item \textbf{Image Encoder}: A streaming strategy is adopted to process each frame individually upon arrival, generating unconditional visual features. A Hiera encoder pretrained with MAE~\cite{he2022masked} is employed to extract multi-scale representations that support downstream decoding.
    
    \item \textbf{Memory Attention}: This module integrates features from the current frame with historical frame representations, predicted masks, and prompt signals. Cross-frame memory association is achieved through a stack of $L$ Transformer layers.
    
    \item \textbf{Prompt Encoder}: Various types of prompts (e.g., clicks, bounding boxes, masks) are encoded and injected into the frame representations to guide target-specific segmentation.
    
    \item \textbf{Mask Decoder}: Composed of two-way Transformer blocks, the decoder jointly updates prompt and visual embeddings, and supports the generation of multiple candidate masks when the prompt is ambiguous.
    
    \item \textbf{Memory Encoder}: Fuses predicted masks with unconditional frame features to extract memory representations, which are subsequently stored in the memory bank.
    
    \item \textbf{Memory Bank}: Two FIFO (first-in, first-out) queues are maintained to store the history of prompted and unprompted frame memories, providing temporal context for future segmentation predictions.
    
\end{itemize}

For detailed implementation, please refer to the original paper~\cite{ravi2024sam}.

\subsection{Manual Inspection in the Data Annotation and Quality Control Process}
\label{sec:appendix_A1}

To construct a high-quality fire segmentation dataset from satellite imagery, we developed a structured annotation pipeline that integrates collaborative labeling with multi-stage quality control. The overall process consists of four key stages: Initial Annotation, Cross-Review, Conflict Resolution, and Final Confirmation.

\paragraph{Initial Annotation.} Annotators delineated fire-affected regions using polygon masks based primarily on infrared imagery, with RGB (visible-spectrum) imagery serving as an auxiliary reference. These annotations were created frame-by-frame using the polygon tool in CVAT (Computer Vision Annotation Tool), as shown in Figure~\ref{fig:a1}.

\paragraph{Cross-Review.} Each annotated mask was independently reviewed by two separate reviewers. An annotation was marked as \textit{approved} only if both reviewers agreed on its spatial accuracy. If either reviewer rejected the mask due to misalignment or imprecise boundaries, the annotation entered the conflict resolution stage.

\paragraph{Conflict Resolution.} In this stage, annotators refined rejected masks using the original imagery and the initial polygon as references. The revised mask was then resubmitted for another round of review. This iterative process continued until both reviewers approved the mask, ensuring inter-annotator agreement and boundary precision.

\paragraph{Final Confirmation.} Once approved, the annotations were finalized and exported for downstream model training. The final outputs included either rasterized binary masks or vector-based polygon representations, depending on the training pipeline requirements.

\paragraph{Implementation Details.} Frame sampling was performed using one of two methods: (1) setting the \textit{Frame Step} parameter during CVAT task creation to import every $N$th frame, or (2) using FFmpeg for offline frame extraction, followed by batch import into CVAT. All annotation and review procedures were conducted within CVAT, ensuring traceability and consistent tooling across stages.

\subsection{Calculation of Mask Annotation Accuracy Metrics}
\label{sec:appendix_A2}

To quantitatively evaluate the segmentation accuracy of masks generated by SAM2 against manually annotated ground truth across five representative regions, we adopt four widely used evaluation metrics: \textbf{accuracy}, \textbf{mean Intersection-over-Union (mIoU)}, \textbf{commission error}, and \textbf{omission error}.

\textit{Accuracy} is defined as the ratio of correctly classified pixels (true positives + true negatives) to the total number of pixels:
\begin{equation}
\mathrm{Accuracy} = \frac{TP + TN}{TP + TN + FP + FN}.
\end{equation}
Values above 0.85 are generally indicative of reliable overall performance.

\textit{mIoU} measures the average overlap between predicted and ground truth regions across all classes (foreground and background), defined as:
\begin{equation}
\mathrm{IoU} = \frac{TP}{TP + FP + FN}, \quad
\mathrm{Mean\ IoU} = \frac{1}{C} \sum_{i=1}^{C} \mathrm{IoU}_i.
\end{equation}
where $C$ denotes the number of classes. A mIoU above 0.5 is typically considered acceptable.

\textit{Commission error} quantifies the proportion of background pixels incorrectly classified as fire (false positives), normalized by the total predicted fire region:
\begin{equation}
\mathrm{Commission\ Error} = \frac{FP}{TP + FP}.
\end{equation}

\textit{Omission error} measures the proportion of fire pixels that were not detected by the model (false negatives), normalized by the total actual fire region:
\begin{equation}
\mathrm{Omission\ Error} = \frac{FN}{TP + FN}.
\end{equation}

Lower values of both error metrics are preferred. Together, these indicators provide a comprehensive assessment of the segmentation quality and robustness of the SAM2 model.

As reported in Table~\ref{tab:accuracy_metrics}, SAM2 demonstrates consistent performance across the five evaluation regions. Accuracy ranges from 0.835 to 0.966, mIoU from 0.623 to 0.782, commission error between 0.045 and 0.101, and omission error from 0.002 to 0.023. These results suggest that SAM2 achieves high segmentation quality with strong generalization across diverse spatial contexts.

\begin{table}[t!]
\centering
\caption{Accuracy metrics of SAM2 across five evaluation regions.}
\setlength{\tabcolsep}{3pt}
\renewcommand{\arraystretch}{1.35}
\begin{tabular}{c|c|c|c|c}
\hline
\textbf{Region} & \textbf{Accuracy $\uparrow$} & \textbf{\makecell{mIoU $\uparrow$}} & \textbf{\makecell{Commission \\ Error $\downarrow$}} & \textbf{\makecell{Omission \\ Error $\downarrow$}} \\ \hline
A & 0.923 & 0.665 & 0.101 & 0.023 \\ 
B & 0.955 & 0.781 & 0.068 & 0.002 \\ 
C& 0.948 & 0.631 & 0.100 & 0.015 \\ 
D & 0.835 & 0.782 & 0.045 & 0.013 \\ 
E & 0.966 & 0.623 & 0.064 & 0.023 \\ \hline
Mean & 0.925 &  0.696 & 0.076 & 0.015 \\ \hline
\end{tabular}
\label{tab:accuracy_metrics}
\end{table}

\section{Additional Information About the New Paradigm}

\subsection{Detailed Derivation of the Diffusion Model}
\label{sec:ddpm}

Let $x_0\in\mathcal{R}^{d}$ be a sample from the data distribution $p_{\mathrm{data}}$. A sample $x_t$ is progressively corrupted through the forward diffusion process (FDP) over time steps $t=0$ to $t=T$, governed by the following transition kernel:
\begin{equation}
    q(x_t \mid x_{t-1}) = \mathcal{N}\left(x_t; \sqrt{1 - \beta_t} \, x_{t-1}, \beta_t \, \mathbf{I} \right)
\label{eq:1}
\end{equation}
Furthermore, $x_t$ can be directly sampled from $x_0$ using the following cumulative kernel:
\begin{equation}
    q_t(x_t|x_0) = \mathcal{N}(x_t; \sqrt{\bar{\alpha}_t}x_0, (1 - \bar{\alpha}_t)I) \implies x_t = \sqrt{\bar{\alpha}_t}x_0 + \sqrt{1 - \bar{\alpha}_t} \boldsymbol{\epsilon},
\label{eq:2}
\end{equation}
where $\bar{\alpha}_t=\Pi_{s=1}^t(1-\beta_s)$, $\boldsymbol{\epsilon}\sim \mathcal{N}(0,I)$.

By reversing the FDP and starting from the Gaussian noise $x_T$, the reverse diffusion process (RDP) can be solved to generate new samples. The RDP can be computed using the following transition kernel~\cite{ho2020denoising,song2020score}:
\begin{equation}
    p_t(x_{t-1} \mid x_t, x_0) = \mathcal{N}(x_{t-1}; \tilde{\mu}_t(x_t, x_0), \tilde{\beta}_t I),
\label{eq:3}
\end{equation}
where, $\tilde{\mu}_t(x_t, x_0) = 
\frac{\sqrt{\bar{\alpha}_{t-1}} \beta_t}{1 - \bar{\alpha}_t} x_0 +
\frac{\sqrt{\alpha_t}(1 - \bar{\alpha}_{t-1})}{1 - \bar{\alpha}_t} x_t$, $\tilde{\beta}_t = \frac{1 - \bar{\alpha}_{t-1}}{1 - \bar{\alpha}_t} \beta_t$.

Since $x_0$ is unknown given $x_t$, it can be estimated using the Equation \ref{eq:2}: $\hat{x}_0 = \frac{x_t - \sqrt{1 - \bar{\alpha}_t} \boldsymbol{\epsilon}}{\sqrt{\bar{\alpha}_t}}$,
where  $\boldsymbol{\epsilon}_\theta(x_t \mid t)$ represents the estimation of the noise $\boldsymbol{\epsilon}$ by a time-conditioned neural network, with parameters $\theta$. This allows us to reverse the process from noise back to data. The loss function for the neural network is:
\begin{equation}
    L(\theta) = \mathbb{E}_{t, x_0 \sim p_{\text{data}}, \epsilon \sim \mathcal{N}(0, \mathbf{I})} 
\left[ \left\| \boldsymbol{\epsilon} - \epsilon_\theta\left( 
\sqrt{\bar{\alpha}_t} x_0 + \sqrt{1 - \bar{\alpha}_t} \boldsymbol{\epsilon} \,\middle|\, t 
\right) \right\|_2^2 \right].
\label{eq:4}
\end{equation}

Note that the estimation of $\boldsymbol{\epsilon}$ is equivalent to estimating a scaled version of the score function of the noise data (i.e., the gradient of the log density):
\begin{equation}
    \nabla_{x_t} \log q_t(x_t \mid x_0) 
= -\frac{1}{1 - \bar{\alpha}_t} ( x_t - \sqrt{\bar{\alpha}_t} x_0 ) 
= -\frac{1}{\sqrt{1 - \bar{\alpha}_t}} \boldsymbol{\epsilon}.
\label{eq:5}
\end{equation}

Thus, data generation through denoising relies on the score function, which can be viewed as a noise-conditional score-based generation method. By applying the score function to the joint distribution of multiple consecutive frames, the diffusion model can be naturally extended to video generation tasks. While this method is effective for unconditional video generation, a systematic solution for other tasks, such as video prediction, is still lacking.

\subsection{The Algorithm of FiReDiff}
\label{appendix_algo}

The specific training and sampling procedures of FiReDiff are provided in Algorithm \ref{algo 1} and Algorithm \ref{algo 2}, respectively.
\begin{algorithm}[t!]
    \caption{Training of FiReDiff}
    \begin{algorithmic}[1]
        \State \textbf{Input:} Past infrared frames $x_0^\mathrm{ob} \in \mathcal{R}^{L \times X \times Y}$, land-vegetation data $ x^\mathrm{LV} \in \mathcal{R}^{X \times Y}$, environmental meteorological data $ x^\mathrm{EM} \in \mathcal{R}^{L}$, target infrared frames $x_0^\mathrm{ta} \in \mathcal{R}^{L' \times X \times Y}$
        \State \textbf{Output:} Trained denoising function $\boldsymbol{\epsilon}_\theta$
        \For{$i = 1$ \textbf{to} $N$}
            \State $t \sim \mathrm{Uniform}(\{1, \dots, T\})$
            \State Sample $\boldsymbol{\epsilon} \sim \mathcal{N}(0, I) \in \mathcal{R}^{L' \times X \times Y}$
            \State Compute noisy targets: 
            $x_t = \sqrt{\bar{\alpha}_t} \, x_0^\mathrm{ta} + \sqrt{1 - \bar{\alpha}_t} \, \boldsymbol{\epsilon}$
            \State Take gradient step using Equation~\ref{eq:6}:
            
            $\nabla_{\theta} \left\lVert \boldsymbol{\epsilon} - \boldsymbol{\epsilon}_\theta(x_t \mid x_0^\mathrm{ob}, x^\mathrm{LV}, x^\mathrm{EM}, t) \right\rVert^2$
        \EndFor
    \end{algorithmic}
    \label{algo 1}
\end{algorithm}

\begin{algorithm}[t!]
    \caption{Prediction with FiReDiff}
    \begin{algorithmic}[1]
        \State \textbf{Input:} Past infrared frames $x_0^\mathrm{ob} \in \mathcal{R}^{L \times X \times Y}$, land-vegetation data $ x^\mathrm{LV} \in \mathcal{R}^{X \times Y}$, environmental meteorological data $ x^\mathrm{EM} \in \mathcal{R}^{L}$, trained denoising function $\boldsymbol{\epsilon}_\theta$, SAM2$(\cdot)$ with prompts
        \State \textbf{Output:} Predicted infrared frames $x_0^\mathrm{ta} \in \mathcal{R}^{L' \times X \times Y}$, predicted masks $m^\mathrm{ta} \in \mathcal{R}^{L' \times X \times Y}$
        \State Sample $x_T^\mathrm{ta} \sim \mathcal{N}(0, I) \in \mathcal{R}^{L' \times X \times Y}$
        \For{$t = T$ \textbf{to} $1$}
            \State Sample $x_{t-1}^\mathrm{ta}$ using Equation~\ref{eq:3}
        \EndFor
        \State $\hat{m}^\mathrm{ta} = \mathrm{SAM2}(x_0^\mathrm{ta}, \text{Prompts})$
    \end{algorithmic}
    \label{algo 2}
\end{algorithm}

\section{Additional Experimental Details}
\subsection{Experimental Configurations}
\label{resolution}
The FireSentry dataset processes video frames through systematic downsampling: spatial resolution is reduced from 1280×1024 pixels to 256×256 (ground sampling distance shifting from 0.5 meter to 2 meter per pixel), while temporal resolution decreases from 30 FPS to 1 FPS. After removing anomalous frames, the curated dataset comprises 20,710 valid frames distributed regionally as follows: region A (6,000 frames), region B (2,135 frames), region C (1,005 frames), region D (3,250 frames), and region E (8,320 frames).
The dataset—comprising video frames paired with corresponding environmental metadata—is partitioned in an 8:1:1 ratio for model training, validation, and testing.

\subsection{Benchmark Models}
\label{app_benchmark}

We conduct a benchmark evaluation of seven representative existing models on the proposed FireSentry dataset and present a preliminary exploration of our proposed paradigm, FiReDiff. All models are categorized into four types: physics-based models, data-driven models, generative models, and FiReDiff-based models. In the implementation, in addition to the past video frames, environmental information is incorporated as a conditional input to guide the denoising process within the diffusion models.

For physics-based models, we utilized the foundational partial differential equations from the WRF-SFIRE simulator to model the propagation of combustion waves ~\cite{mandel2011coupled}.

For data-driven models, we evaluate Earthformer ~\cite{gao2022earthformer} and PredRNN ~\cite{wang2022predrnn}. Earthformer leverages an efficient and flexible Cuboid Attention mechanism for spatio-temporal prediction. PredRNN introduces a novel spatio-temporal LSTM unit that jointly models spatial and temporal dependencies to generate future frames.

For generative models, we select MCVD ~\cite{voleti2022mcvd}, STDiff ~\cite{ye2024stdiff}, VDT ~\cite{lu2023vdt}, and DynamiCrafter ~\cite{xing2024dynamicrafter}. MCVD is a denoising diffusion model guided by conditional likelihood scoring, using independently masked past frames for video generation. STDiff first models motion dynamics using a neural stochastic differential equation and then generates future frames via a conditional diffusion process. VDT introduces a Transformer architecture into diffusion-based video generation and proposes a unified spatio-temporal masking mechanism to accommodate diverse video generation scenarios. DynamiCrafter is a text-to-video diffusion model that incorporates motion priors to condition generation on image content.

For the FiReDiff-based models, we adapt the four generative models above—MCVD, STDiff, VDT, and DynamiCrafter—to construct MCVD*, STDiff*, VDT*, and DynamiCrafter* under our proposed paradigm. To evaluate the performance of FiReDiff on the fire video forecasting task, these models first predict future frames in the infrared modality, and the predicted infrared videos are then processed by the SAM2 model to obtain corresponding segmentation masks, ultimately producing the final fire mask videos.

\noindent\textbf{Key Innovation}: 
While conventional approaches (physics-based, data-driven, and generative models) directly predict mask sequences, FiReDiff establishes the first cross-modal framework that decouples infrared forecasting from mask generation—leveraging complementary modality strengths for enhanced wildfire modeling.

\subsection{Evaluation Metrics}
\label{app_meric}

To evaluate both video generation quality and fire mask segmentation accuracy, we adopt a total of eight metrics. Specifically, we use four standard metrics for assessing video generation: \textbf{PSNR}, \textbf{SSIM}, \textbf{LPIPS}, and \textbf{FVD}; and four metrics for evaluating fire mask segmentation: \textbf{AUPRC}, \textbf{F1 Score}, \textbf{IoU}, and \textbf{MSE}. Definitions are provided below for each metric.

\textit{PSNR (Peak Signal-to-Noise Ratio)} quantifies the fidelity between predicted and ground-truth video frames. Higher values indicate better reconstruction quality:
\begin{equation}
\mathrm{PSNR} = 10 \cdot \log_{10} \left( \frac{\mathrm{MAX}^2}{\mathrm{MSE}} \right)
\end{equation}
where $\mathrm{MAX}$ is the maximum possible pixel value (e.g., 255), and $\mathrm{MSE}$ denotes mean squared error between corresponding frames.

\begin{figure}[t!]
    \centering
    \includegraphics[width=0.45\textwidth]{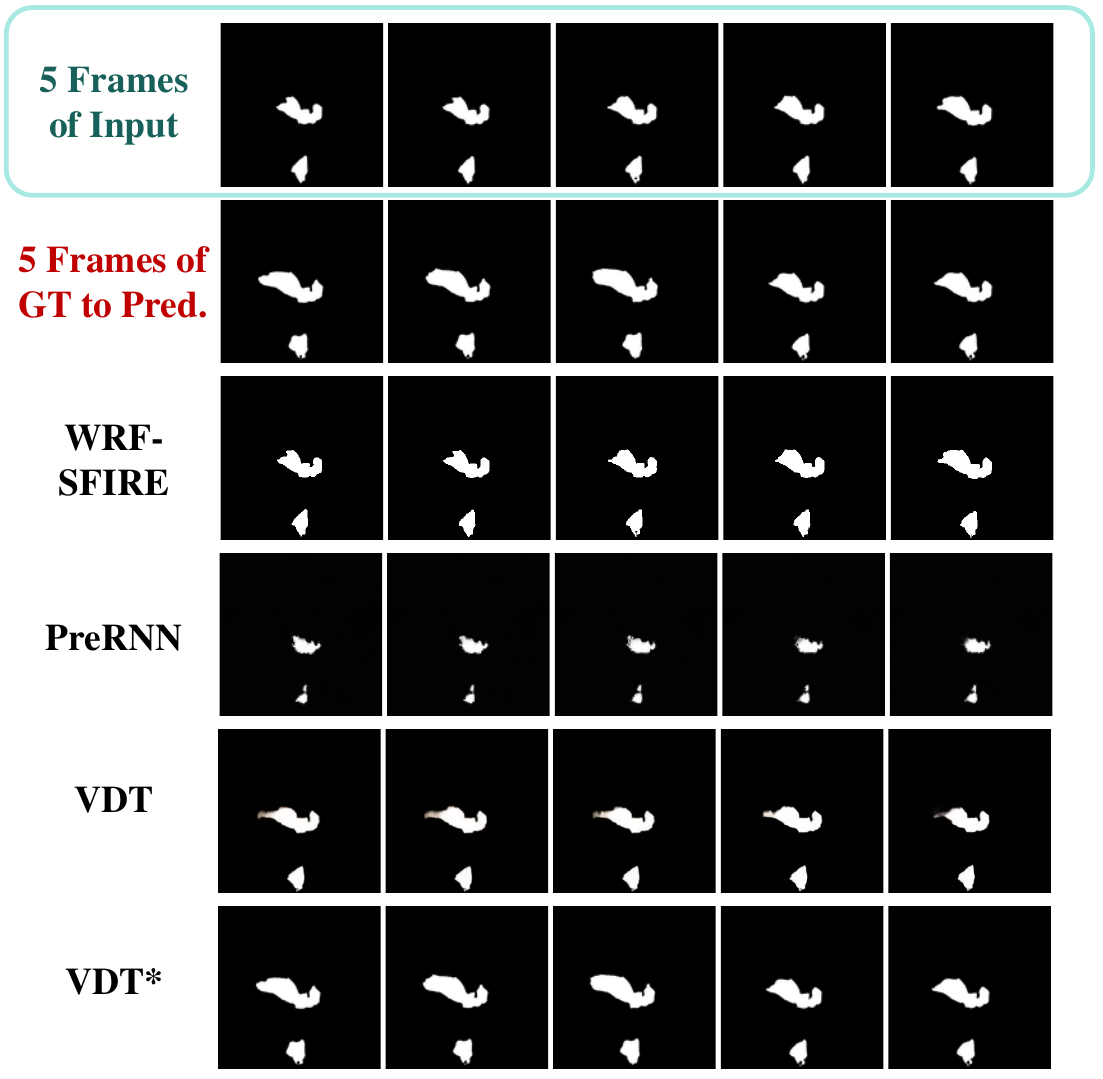}
    \caption{The visualization of fire mask results.}
    \label{fig:visual_mask}
\end{figure}

\begin{figure}[t!]
    \centering
    \includegraphics[width=0.45\textwidth]{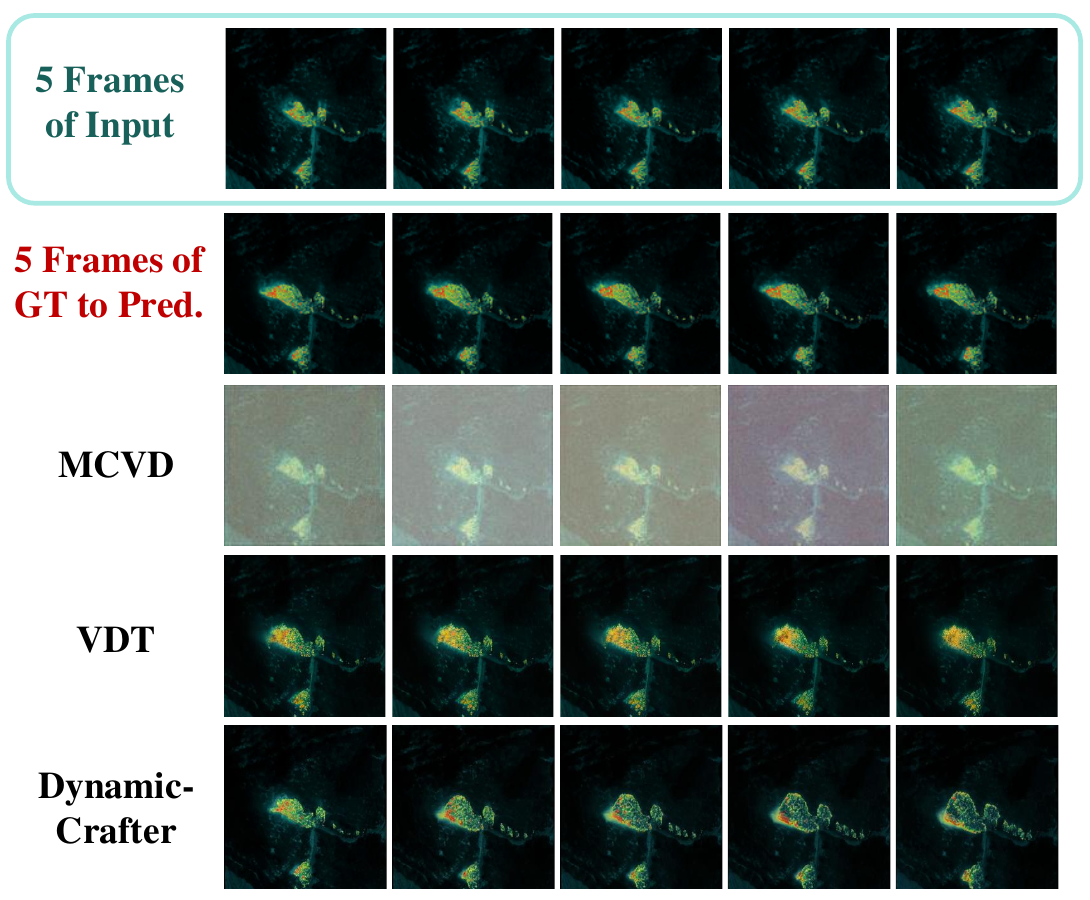}
    \caption{The visualization of fire infrared prediction results.}
    \label{fig:visual_infra}
\end{figure}

\textit{SSIM (Structural Similarity Index)} evaluates perceptual similarity between frames, taking into account luminance, contrast, and structural information. Higher scores imply better perceptual quality:
\begin{equation}
\mathrm{SSIM}(x, y) = \frac{(2\mu_x\mu_y + C_1)(2\sigma_{xy} + C_2)}{(\mu_x^2 + \mu_y^2 + C_1)(\sigma_x^2 + \sigma_y^2 + C_2)}
\end{equation}
where $\mu_x$, $\mu_y$, $\sigma_x^2$, $\sigma_y^2$, and $\sigma_{xy}$ represent the means, variances, and covariance of images $x$ and $y$, respectively. $C_1$ and $C_2$ are small constants to avoid division by zero.


\textit{LPIPS (Learned Perceptual Image Patch Similarity)} measures perceptual similarity between images by comparing deep feature activations extracted from a pretrained neural network. Lower LPIPS values indicate higher perceptual similarity. Formally, given two images \(x\) and \(\hat{x}\), LPIPS is computed as
\begin{equation}
\mathrm{LPIPS}(x, \hat{x}) = \sum_{l} \frac{1}{H_l W_l} \sum_{h=1}^{H_l} \sum_{w=1}^{W_l} w_l \cdot \left\| \hat{y}^l_{h,w} - y^l_{h,w} \right\|_2^2,
\end{equation}
where \(y^l\) and \(\hat{y}^l\) are the feature maps at layer \(l\) of the network for images \(x\) and \(\hat{x}\), respectively; \(H_l, W_l\) denote the spatial dimensions of the feature map; and \(w_l\) are learned scalar weights for each layer. Since LPIPS relies on a pretrained network, it does not have a closed-form expression.

\textit{FVD (Fréchet Video Distance)} measures the distributional distance between generated and real video feature distributions. Lower values reflect better realism:
\begin{equation}
\mathrm{FVD} = \|\mu_r - \mu_g\|^2 + \mathrm{Tr}\left( \Sigma_r + \Sigma_g - 2(\Sigma_r \Sigma_g)^{1/2} \right)
\end{equation}
where $(\mu_r, \Sigma_r)$ and $(\mu_g, \Sigma_g)$ are the means and covariances of feature representations from real and generated videos.

\textit{AUPRC (Area Under the Precision-Recall Curve)} evaluates segmentation performance by summarizing the trade-off between precision and recall over varying classification thresholds. A higher AUPRC value indicates better overall segmentation quality. 
Formally, given precision \( P(r) \) as a function of recall \( r \), the AUPRC is defined as the integral:
\begin{equation}
\mathrm{AUPRC} = \int_0^1 P(r) \, dr,
\end{equation}
where precision and recall are computed as
\begin{equation}
\text{Precision} = \frac{TP}{TP + FP}, \quad \text{Recall} = \frac{TP}{TP + FN},
\end{equation}
with \(TP\), \(FP\), and \(FN\) denoting true positives, false positives, and false negatives, respectively.

\textit{F1 Score} captures the harmonic mean of precision and recall, providing a balanced measure of accuracy:
\begin{equation}
\mathrm{F1\ Score} = \frac{2 \cdot \mathrm{Precision} \cdot \mathrm{Recall}}{\mathrm{Precision} + \mathrm{Recall}}
\end{equation}
Higher values represent more accurate fire mask predictions.

\textit{IoU (Intersection over Union)} measures the overlap between predicted and ground-truth segmentation masks:
\begin{equation}
\mathrm{IoU} = \frac{|P \cap G|}{|P \cup G|}
\end{equation}
where $P$ and $G$ are the predicted and ground-truth mask regions, respectively. Larger values indicate better alignment.

\textit{MSE (Mean Squared Error)} quantifies the average squared difference between predicted and ground-truth mask pixels:
\begin{equation}
\mathrm{MSE} = \frac{1}{N} \sum_{i=1}^{N} (x_i - y_i)^2
\end{equation}
where $x_i$ and $y_i$ denote pixel values of the predicted and ground-truth masks. Lower values indicate more accurate segmentation.

\subsection{Visualizations of the Mask Video Prediction Results}
\label{vis_mask}

Figure~\ref{fig:visual_mask} presents visual comparisons of fire mask predictions across different model types. Among them, the VDT\* model, operating under the FiReDiff paradigm, delivers the most accurate and temporally coherent predictions. Notably, compared to its original version that predicts solely within the mask modality, the FiReDiff-enhanced VDT\* exhibits substantial performance improvements, highlighting the advantages of cross-modal prediction.

\begin{table}[t!]
\centering
\caption{Model performance comparison on infrared video prediction.}
\setlength{\tabcolsep}{2.5pt}
\begin{tabular}{c|c|cccc}
\Xhline{2\arrayrulewidth}
\multicolumn{2}{c|}{} & \multicolumn{4}{c}{\textbf{Video Prediction Metrics}} \\ 
\cmidrule(lr){3-6}
\multicolumn{2}{c|}{\multirow{-2}{*}{\centering\textbf{Method}}} & \textbf{PSNR↑} & \textbf{SSIM↑} & \textbf{LPIPS↓} & \textbf{FVD↓} \\ 
\Xhline{2\arrayrulewidth}
\multirow{4}{*}{\textbf{\centering\makecell{FiReDiff \\ Paradigm}}} & MCVD* & 7.477 & 0.129 & 0.711 & 636.087 \\ 
& STDiff* & 14.019 & 0.415 & 0.184 & 507.76 \\ 
& VDT* & 18.128 & 0.701 & 0.091 & 198.166 \\ 
& DynamiCrafter* & 14.548 & 0.675 & 0.183 & 596.407 \\
\Xhline{2\arrayrulewidth}
\end{tabular}
\label{tab:infra pred}
\end{table}

\begin{table}[t!]
\centering
\caption{Effect of different pixel‐error tolerances on fire detection accuracy.}
\renewcommand{\arraystretch}{1.3}
\setlength{\tabcolsep}{4.8pt}
\begin{adjustbox}{width=0.45\textwidth}
\begin{tabular}{c|cccc}
\Xhline{2\arrayrulewidth}
{\textbf{Pixel Error}} & 3 (1.5 m)   & 4 (2 m)   & 5 (2.5 m)   & 6 (3 m)   \\
\Xhline{2\arrayrulewidth}
{Fire Points Accuracy}     & 87.70\%         & 92.40\%       & 95.20\%         & 96.50\%       \\
{Fire Lines Accuracy}     & 80.30\%         & 87.10\%       & 91.00\%         & 93.10\%       \\
\Xhline{2\arrayrulewidth}
\end{tabular}
\end{adjustbox}
\label{fire_acc}
\end{table}

\begin{figure}[t!]
    \centering
    \includegraphics[width=0.45\textwidth]{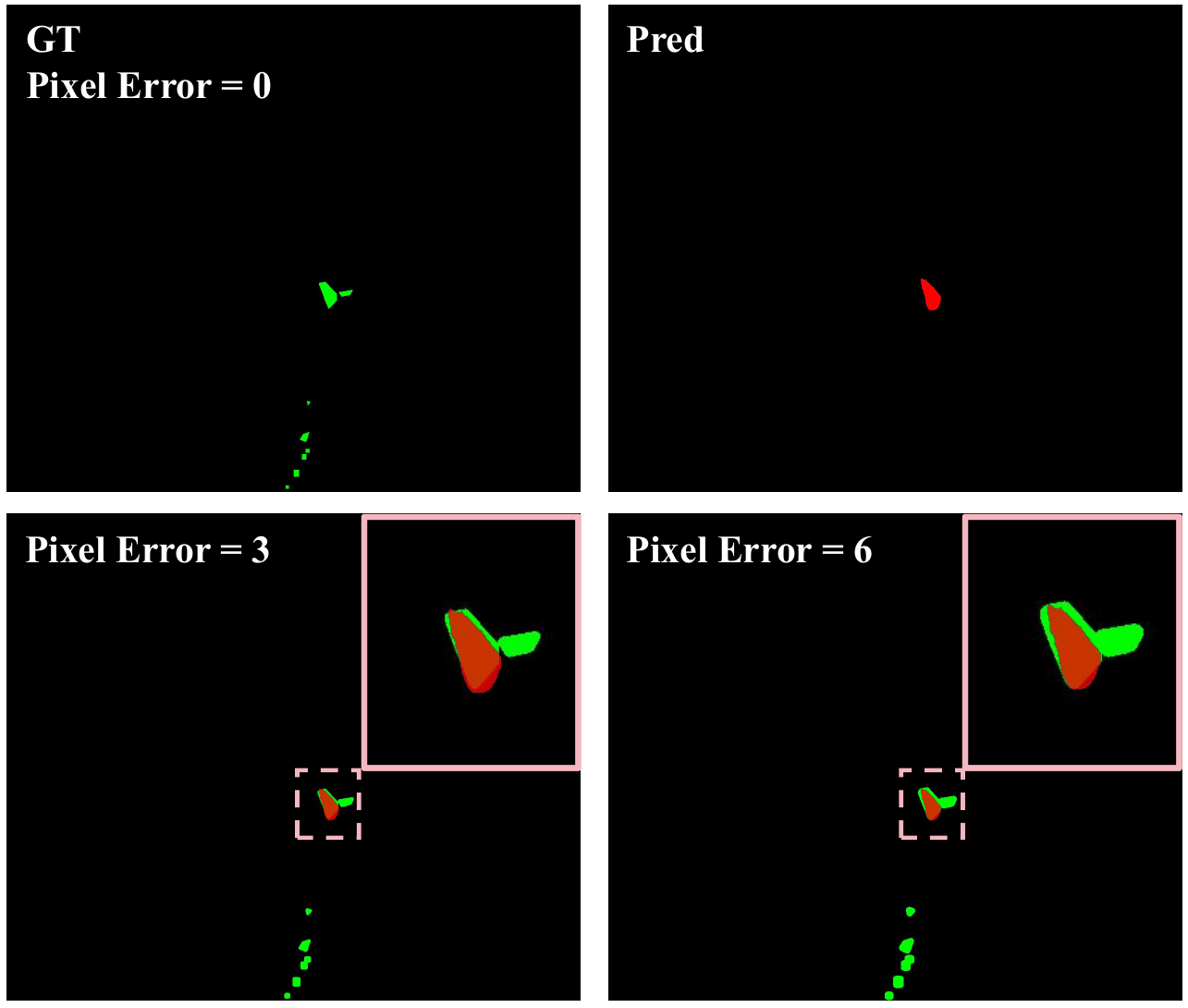}
    \caption{The visualization of fire points recognition.}
    \label{fire_point}
\end{figure}

\begin{figure}[t!]
    \centering
    \includegraphics[width=0.45\textwidth]{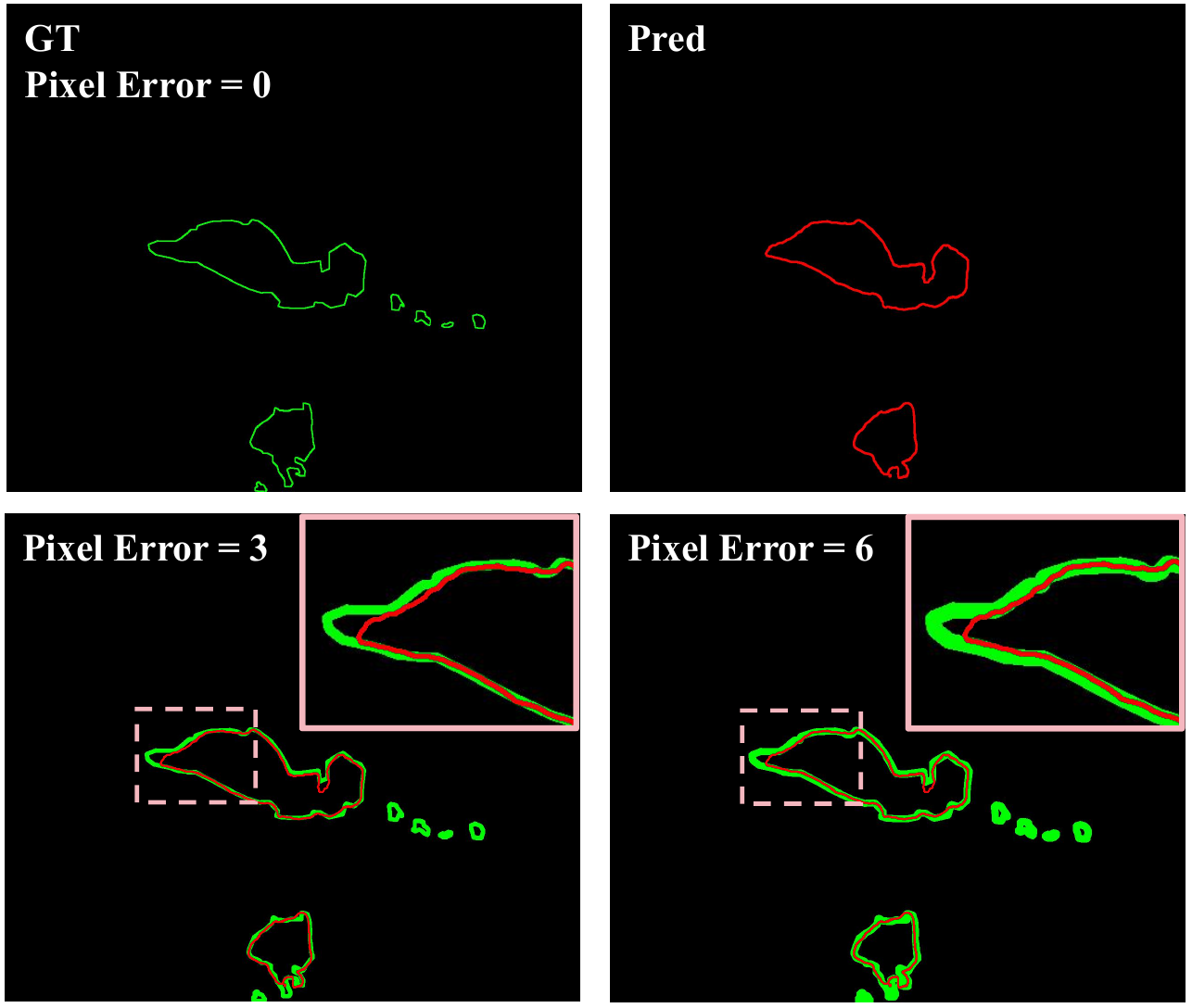}
    \caption{The visualization of fire lines recognition.}
    \label{fire_line}
\end{figure}

\subsection{Results of Infrared Video Prediction Using FiReDiff}
\label{vis_infra}

The evaluation metrics and visualizations for infrared video prediction under the FiReDiff paradigm are presented in Table~\ref{tab:infra pred} and Figure~\ref{fig:visual_infra}, respectively. Among the models compared, the Transformer-based VDT achieves the highest performance across all four evaluation metrics, indicating superior video generation quality. In contrast, the U-Net-based MCVD exhibits suboptimal denoising capabilities, resulting in lower-quality video outputs. These observations highlight the advantages of Transformer architectures in modeling complex spatio-temporal dynamics for infrared video prediction tasks.

\begin{table*}[t!]
\centering
\caption{Performance comparison under spatial resolution scaling. 
Percentages in parentheses denote the performance drop when resolution increases from 256×256 to 512×512. Abbreviations: DC refers to DynamiCrafter; DC* refers to DynamiCrafter*; Res. refers to resolution.}
\renewcommand{\arraystretch}{1.3}
\setlength{\tabcolsep}{1pt}
\begin{adjustbox}{width=\textwidth}
\begin{tabular}{c|cccc|cccc}
\Xhline{2\arrayrulewidth}
\textbf{Res.} & \multicolumn{4}{c|}{\textbf{Video Prediction Metrics}} & \multicolumn{4}{c}{\textbf{Fire Mask Metrics}} \\
\cmidrule(lr){2-5} \cmidrule(lr){6-9}
 & \textbf{PSNR↑} & \textbf{SSIM↑} & \textbf{LPIPS↓} & \textbf{FVD↓} & \textbf{AUPRC↑} & \textbf{F1 score↑} & \textbf{IoU↑} & \textbf{MSE↓} \\
\Xhline{2\arrayrulewidth}
\textbf{DC-256 Res.} & 7.415 & 0.729 & 0.252 & 1451.692 & 0.946 & 0.259 & 0.483 & 0.230 \\
\textbf{DC*-256 Res.} & 10.324 & 0.815 & 0.198 & 1024.364 & 0.974 & 0.412 & 0.451 & 0.154 \\
\hline
\textbf{DC-512 Res.} & 3.827(-48.4\%) & 0.379(-48.0\%) & 0.487(-93.2\%) & 2512.314(-73.1\%) & 0.485(-48.7\%) & 0.133(-48.6\%) & 0.251(-48.0\%) & 0.420(-82.6\%) \\
\textbf{DC*-512 Res.} & 7.771(-24.7\%) & 0.654(-19.8\%) & 0.274(-38.4\%) & 1436.546(-40.2\%) & 0.784(-19.5\%) & 0.301(-26.9\%) & 0.359(-20.4\%) & 0.201(-30.5\%) \\
\Xhline{2\arrayrulewidth}
\end{tabular}
\end{adjustbox}
\label{tab:spatial_scaling}
\end{table*}

\subsection{Recognition Rates of Fire Ignition Points and Fire Lines}
\label{fire_line_point}

Fire lines denote the boundaries of burned regions, whereas fire points correspond to the initial ignition locations. We compute the detection accuracy as follows:
\begin{align}
\hat{G} &= G \oplus B_{r}, \\[4pt]
\mathrm{Accuracy} &= \frac{\lvert P \cap \hat{G}\rvert}{\lvert \hat{G}\rvert} \times 100\%,
\end{align}
where $G$ is the ground‑truth binary mask (burned region = 1, background = 0), $P$ is the predicted binary mask, and $\oplus B_{e}$ denotes dilation with a structuring element of radius $e$. The operator $\lvert\cdot\rvert$ returns the number of pixels in a set. 

Figure~\ref{fire_point} and~\ref{fire_line} illustrate visual comparisons between detected and ground‑truth fire points and fire lines under varying tolerance radius $e$. Each pixel error translates to a 0.5 meter real‑world deviation. As summarized in Table~\ref{fire_acc}, when the allowable error radius is 1.5 meter, fire‑point accuracy reaches 87.7\% and fire‑line accuracy reaches 80.3\%. Increasing the tolerance to 3 meter raises fire‑point accuracy to 96.5\% and fire‑line accuracy to 93.1\%.

\subsection{Spatial Resolution Scaling Analysis}
\label{spatial_exp}
To further validate the spatial stability of the FiReDiff paradigm, we compare DynamiCrafter with DynamiCrafter* (which applies FiReDiff) at resolutions of 256×256 and 512×512. As shown in Table~\ref{tab:spatial_scaling}, when spatial resolution doubles, FiReDiff significantly mitigates performance degradation and improves model robustness by 45.4\% on average.

\end{document}